\documentclass[sigconf]{acmart}

\usepackage{amsmath,amsfonts,bm}

\def\eqref#1{equation~\ref{#1}}
\def\1{\bm{1}}

\DeclareMathAlphabet{\mathsfit}{\encodingdefault}{\sfdefault}{m}{sl}
\SetMathAlphabet{\mathsfit}{bold}{\encodingdefault}{\sfdefault}{bx}{n}

\usepackage[utf8]{inputenc} % allow utf-8 input
\usepackage[T1]{fontenc}    % use 8-bit T1 fonts
\usepackage{hyperref}       % hyperlinks
\usepackage{url}            % simple URL typesetting
\usepackage{booktabs}       % professional-quality tables
\usepackage{amsfonts}       % blackboard math symbols
\usepackage{nicefrac}       % compact symbols for 1/2, etc.
\usepackage{microtype}      % microtypography
\usepackage{xcolor}         % colors
\usepackage{graphicx}
\usepackage{amsmath}
\usepackage{multirow}
\usepackage{tabularx}
\usepackage{bibunits}
\defaultbibliography{ref}
\defaultbibliographystyle{unsrt}

\usepackage{tikz}
\usetikzlibrary{patterns}
\usetikzlibrary{arrows, automata}

\usepackage{caption,subcaption}
\usepackage{physics}
\usepackage{mathtools}
\usepackage{enumerate}% http://ctan.org/pkg/enumerate
\usepackage{wrapfig}

\usepackage{colortbl}

\usepackage{pdfrender}
\newcommand\tablebf[1]{\textpdfrender{TextRenderingMode=FillStroke,LineWidth=0.6}{#1}} % The LineWidth is the boldness

\usepackage{enumitem}
\newlist{todolist}{itemize}{2}
\setlist[todolist]{label=$\square$}
\usepackage{pifont}
\AtBeginDocument{%
  }

\copyrightyear{2026}
\acmYear{2026}
\setcopyright{cc}
\setcctype{by-nc-nd}

\acmConference[KDD '26]{Proceedings of the 32nd ACM SIGKDD Conference on Knowledge Discovery and Data Mining V.2}{August 09--13, 2026}{Jeju Island, Republic of Korea}
\acmBooktitle{Proceedings of the 32nd ACM SIGKDD Conference on Knowledge Discovery and Data Mining V.2 (KDD '26), August 09--13, 2026, Jeju Island, Republic of Korea}
\acmDOI{10.1145/3770855.3817519}
\acmISBN{979-8-4007-2259-2/2026/08}
\begin{document}

\title{Causality for Tabular Data Synthesis:
A High-Order Structure Causal Benchmark Framework}

% Equal contribution note (applied to both first authors via authornotemark)
% ---------------- Authors ----------------

% -------- Authors (ACM compliant, stack-safe) --------

\author{Zineb Senane}
\orcid{0009-0001-6451-0136}
\affiliation{%
  \institution{KTH Royal Institute of Technology}
  \city{Stockholm}
  \country{Sweden}}
% One (1) global note to avoid maketitle recursion/stack overflow
\authornote{
Equal first authors: Zineb Senane and Axel Karlsson.
Cheng Zhang’s contribution was mainly during Microsoft Research, Cambridge, UK; currently at ELLIS and EIT.
Ruibo Tu’s contribution was mainly during his time at KTH; he is currently also affiliated with Qlik, Stockholm, Sweden. Zineb Senane's contribution was mainly during KTH, currently affiliated with Fever Energy, Stockholm, Sweden.
Corresponding authors: Ruibo Tu (\href{mailto:ruibo@kth.se}{ruibo@kth.se}) and Lele Cao (\href{mailto:lelecao@microsoft.com}{lelecao@microsoft.com}).}

\author{Axel Karlsson}
\authornotemark[1]
\orcid{0000-0002-3089-5525}
\affiliation{%
  \institution{King AI Labs, Microsoft}
  \city{Stockholm}
  \country{Sweden}}

\author{Lele Cao}
\orcid{0000-0002-5680-9031}
\affiliation{%
  \institution{King AI Labs, Microsoft}
  \city{Stockholm}
  \country{Sweden}}

\author{Oleg Smirnov}
\orcid{0000-0003-1433-7037}
\affiliation{%
  \institution{King AI Labs, Microsoft}
  \city{Stockholm}
  \country{Sweden}}

\author{Cheng Zhang}
\orcid{0000-0002-8640-9370}
\affiliation{%
  \institution{ELLIS}
  \city{Cambridge}
  \country{UK}}

\author{Sahar Asadi}
\orcid{0000-0002-4017-3550}
\affiliation{%
  \institution{King AI Labs, Microsoft}
  \city{Stockholm}
  \country{Sweden}}

\author{Hedvig Kjellström}
\orcid{0000-0002-5750-9655}
\affiliation{%
  \institution{KTH Royal Institute of Technology}
  \city{Stockholm}
  \country{Sweden}}

\author{Gustav Eje Henter}
\orcid{0000-0002-1643-1054}
\affiliation{%
  \institution{KTH Royal Institute of Technology}
  \city{Stockholm}
  \country{Sweden}}

\author{Ruibo Tu}
\orcid{0000-0003-1356-9653}
\affiliation{%
  \institution{KTH Royal Institute of Technology}
  \city{Stockholm}
  \country{Sweden}}

\renewcommand{\shortauthors}{Senane et al.}

\begin{abstract}
  Existing evaluations of tabular synthesis models rely primarily on low-order statistics and downstream task performance, leaving multivariate causal relationships that go beyond pairwise correlations largely unmeasured. We argue that a systematic evaluation on high-order structural information is a crucial first step in addressing this issue in tabular data synthesis. In this paper, we present high-order structural causal information as a natural form of prior knowledge and introduce a benchmark framework to evaluate tabular synthesis models. This framework allows us to generate benchmark datasets through a flexible range of data generation processes, allowing for the training of tabular synthesis models using these datasets for further evaluation. We propose multiple benchmark tasks, high-order metrics, and causal inference tasks as downstream tasks for evaluating the quality of synthetic data generated by the trained models. Our experiments demonstrate the effectiveness of the benchmark framework in evaluating the model's ability to capture high-order structural causal information. Furthermore, our benchmarking results provide an initial assessment of state-of-the-art tabular synthesis models. These results reveal significant gaps between ideal and actual performance and highlight how baseline methods differ. We position the framework as a controlled diagnostic benchmark for causal fidelity, complementing existing low-order and downstream evaluations. We open source the benchmark framework, including both code and data along with documentation, to support further research in this area. 
\end{abstract}





\begin{teaserfigure}
  \hbox to \textwidth{\hfil\small
  \textbf{Code: }
  \url{https://github.com/TURuibo/CauTabBench} \quad\
  \textbf{Data: }
  \url{https://doi.org/10.7910/DVN/EB0KCO}
  \hfil}
  \vspace{10pt}
\end{teaserfigure}

\maketitle

\section{Introduction}\label{sec:introduction}

Tabular data are widely used in both industry and natural sciences, yet tabular data remain underexplored in machine learning research~\citep{van2024tabular}. Among the various tasks in the tabular domain, data synthesis is particularly important due to its many applications, such as data augmentation to mitigate data scarcity~\citep{choi2017generating}, pretraining for downstream tasks~\citep{hollmann2023tabpfn}, and privacy preservation~\citep{hernandez2022synthetic}. Recently, the quality of synthetic tabular data has significantly improved with deep diffusion models (DFMs)~\citep{ho2020denoising} and large language models (LLMs)~\citep{brown2020language}. 

Nevertheless, tabular data synthesis continues to face several challenges. These challenges fall into three broad categories.
\textbf{(C1)} Practical issues, such as mixed data types~\citep{ma2020vaem} and missing data.
\textbf{(C2)} Capturing structural information inherent to tabular data, particularly high-order instance and feature dependencies~\citep{li2023graph}, where ``high-order'' refers to multivariate structural causal information beyond pairwise relationships, typically represented through causal graphs or skeletons.
\textbf{(C3)} Synthesizing data in a cross-table context, such as capturing dependencies across tables~\citep{scetbon2024fip}.
This paper focuses on (C2): whether current tabular synthesis models capture high-order causal structure beyond pairwise dependencies.

While many studies have focused on addressing practical issues~\citep{kotelnikov2023tabddpm,kim2022stasy,lee2023codi,zhang2023mixed} that are necessary steps for training synthesis models, fewer have addressed high-order information, which is crucial for complex real-world applications such as in-context prediction~\citep{zhu2023tabular,hollmann2023tabpfn} and generalization across multiple tables~\citep{wang2022transtab,zhu2023xtab}.

One reason for the lack of studies may be the absence of a systematic evaluation of synthesis models on high-order information. This absence not only creates a limited and misleading impression of the performance of synthesis models, but also impedes the development and application of high-order information-aware synthesis models. The evaluation of tabular synthesis models is an active research area. Currently, evaluations primarily focus on the performance of using synthetic data for downstream tasks, known as \emph{extrinsic evaluation}~\citep{bommasani2021opportunities}. Extrinsic evaluation offers a limited understanding of tabular synthesis models restricted by the downstream tasks. In contrast, \emph{intrinsic evaluation} directly evaluates the quality of synthetic data using metrics derived from lower-order statistics such as pairwise correlation-based scores. Intrinsic evaluation with high-order metrics is challenging, as it depends on informative prior knowledge that remains underexplored in the tabular domain~\citep{van2024tabular}.

\begin{figure*}
%\vspace{-10pt}
\centering
\begin{subfigure}{0.7\textwidth}
    \includegraphics[width=\textwidth]{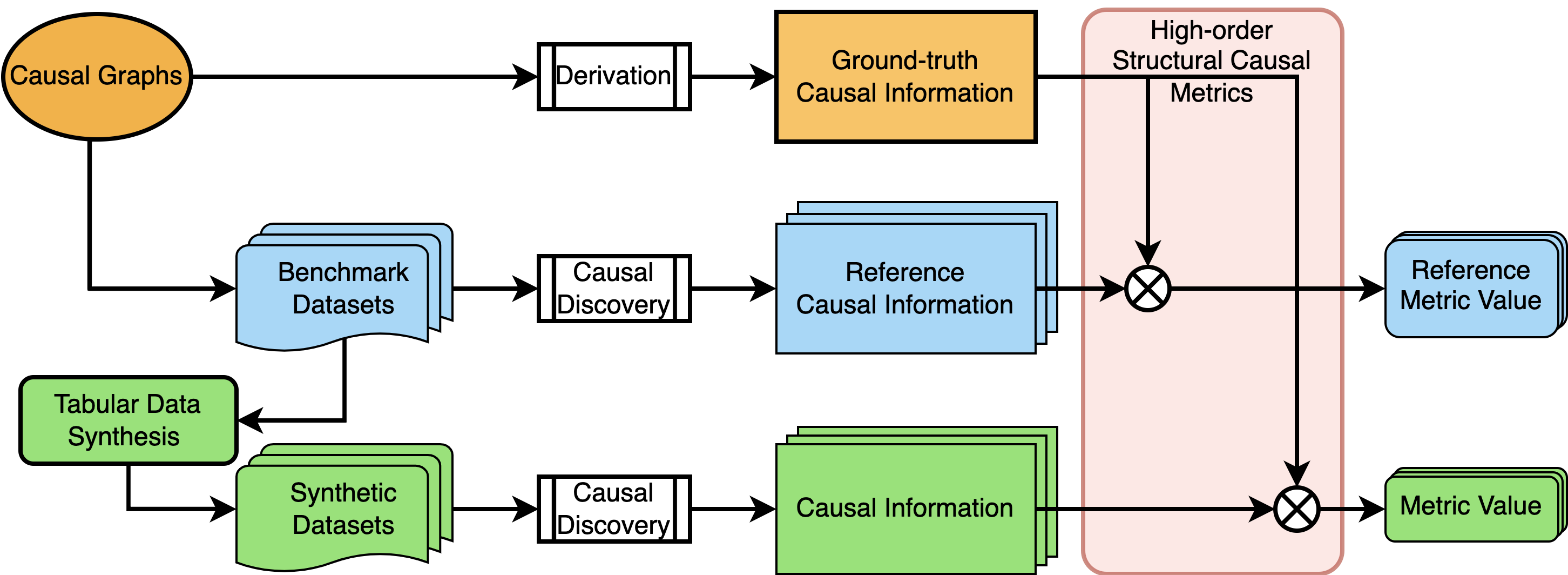}  
    \caption{Benchmark framework.}
    \label{fig:framework}
\end{subfigure}
\begin{subfigure}{0.28\textwidth}
    \includegraphics[width=\textwidth]{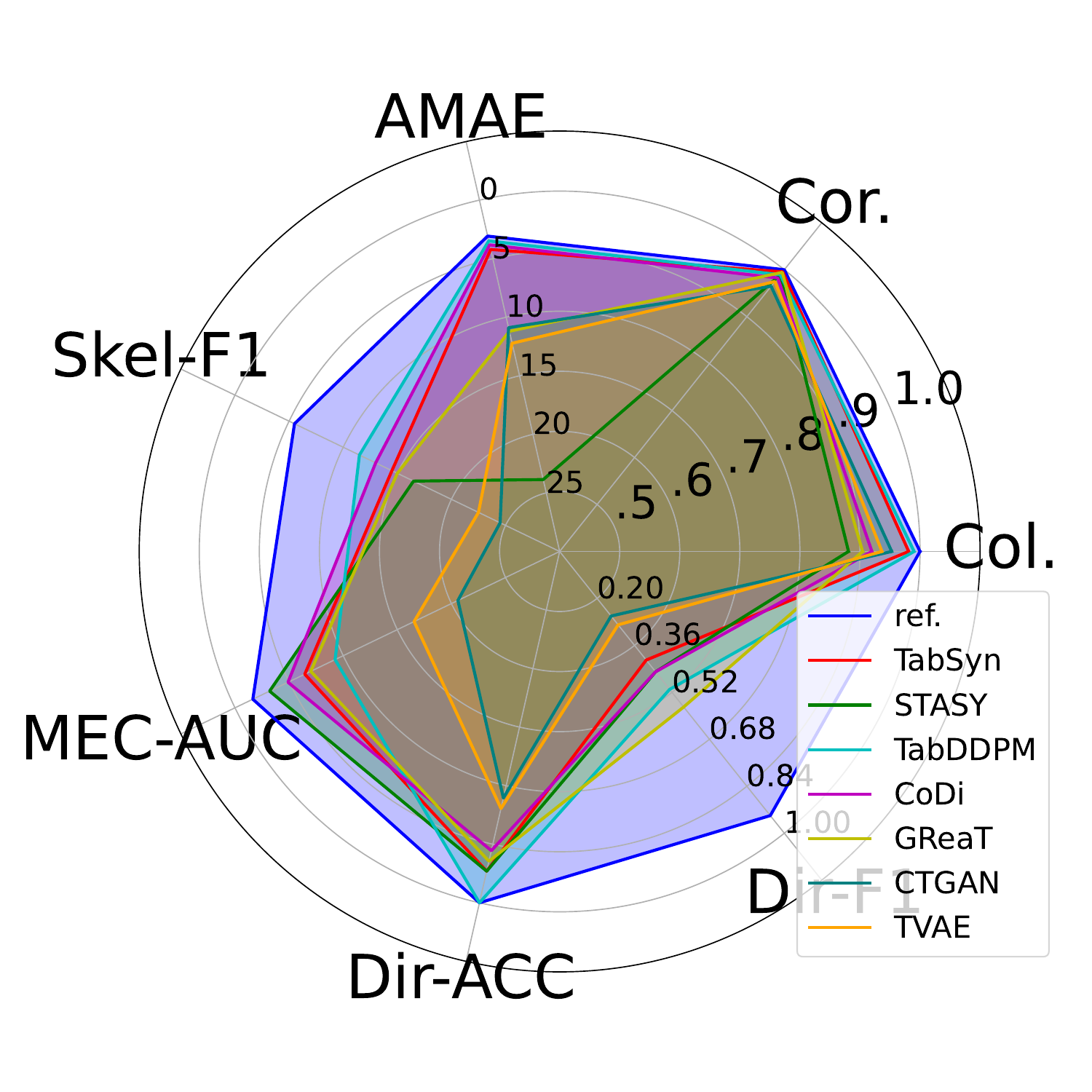}
    \caption{Benchmark results.}\label{fig:lu_radar}
\end{subfigure}
\vspace{-5pt}
\caption{High-order structural causal benchmark framework and results.
Benchmark datasets are generated from sampled causal graphs and used to train tabular synthesis models. Causal information is extracted from both benchmark and synthetic datasets using causal discovery and compared to the ground truth using high-order structural causal metrics. The results highlight model performance across multiple metrics, using benchmark-derived values as references.}\label{fig:overview}
\vspace{-8pt}
\end{figure*}

To benchmark models on high-order structural information, we propose to leverage the study of causal graphical models~\citep{spirtes2000causation,peters2017elements}. These models apply causal prior knowledge for a wide range of machine learning applications~\citep{pearl2016causal,scholkopf2021toward}, yet they are still underexplored in the tabular domain. We view causal graphs as a natural and compact representation of high-order structural information about causal dependencies in tabular data. A few works~\citep{choi2020learning,liu2022goggle,yan2023t2g} have attempted to use general graph properties, such as adjacency matrices and directed acyclicity, for representation learning and tabular data synthesis, but they have all overlooked high-order structural causal information. 
In contrast, our aim is to establish a foundation that covers the essential aspects required for benchmarking tabular synthesis models on high-order structural causal information. We also demonstrate how to utilize benchmarking results to guide future improvements in tabular synthesis models. Specifically, we

\begin{enumerate}[leftmargin=*, topsep=0pt, partopsep=0pt, itemsep=0pt]
    \item introduce high-order structural causal information as prior knowledge for modeling dependencies among the variables of interest in tabular data. We characterize the information into three levels for benchmarking tabular synthesis models (Section~\ref{sec:preliminaries}).
    \item propose a benchmark framework for evaluating tabular synthesis models on high-order structural causal information, which is summarized in Figure~\ref{fig:overview}. 
    Specifically, we illustrate how to generate benchmark datasets and how high-order structural causal tasks and downstream causal-inference tasks can be used for evaluation  (Section~\ref{sec:benchmark}).
    \item demonstrate using our framework to evaluate state-of-the-art DFM and Transformer-based tabular synthesis models on benchmark and real-world datasets (Section~\ref{sec:experiments} and Appendix~\ref{app:exp_details}).
    Our experimental results show a clear gap between the ideal and actual performance of the baseline methods, and strong performance on low-order metrics does not guarantee good performance on high-order causal metrics. These results highlight their shortcomings from various perspectives. 
\end{enumerate}

\section{Related Work}
\label{sec:relatedworks}
To understand tabular data models from different perspectives and to make progress towards better real-world performance, a suite of benchmarks with different purposes is needed. In prior benchmarking efforts,~\cite{grinsztajn2022tree} uses diverse tabular datasets to investigate the performance of tree-based methods; \cite{bommert2020benchmark},~\cite{passemiers2023good} and~\cite{cherepanova2023performance} contribute benchmark datasets to the studies of feature selection;~\cite{malinin2021shifts} and~\cite{gardner2023benchmarking} focus on the robustness to distribution shifts. Moreover, to bridge the gap between real-world applications and synthetic data, \cite{jesus2022turning} provides a larger-scale tabular dataset in finance, including practical challenges. In contrast, SynthCity~\citep{qian2023synthcity} provides a number of synthetic data generators for better model development by avoiding practical issues of real-world data, such as selection bias and missing value issues. Furthermore, in the context of tabular data synthesis,~\cite{hansen2023reimagining} introduces data-centric AI techniques that can provide data profiles, and then propose an evaluation framework to show the importance of integrating data profiles into synthesis models. Unlike previous efforts, our work provides a new perspective for understanding and improving tabular synthesis models by benchmarking on high-order structural causal information.
Related to this, recent work has also discussed the challenges of causal benchmarking and the risk of overly simplistic simulated DAGs~\citep{reisach2021beware}.

Tabular data synthesis was initially based on classical deep generative models~\citep{jordon2018pate,xu2019modeling,morales2022simultaneous}, but has recently flourished with Transformer-based, LLM-based, and DFM-based methods. For example, to generate synthetic tabular data, \citep{gulati2024tabmt} applies a masked Transformer~\citep{vaswani2017attention}; \emph{GReaT}~\citep{borisov2022deep} formulates tabular data as sentences and finetunes a Generative Pre-trained Transformer (GPT-2)~\citep{radford2019language}; \emph{TabDDPM}~\citep{kotelnikov2023tabddpm}, and \emph{CoDi}~\citep{lee2023codi} apply denoising diffusion probabilistic models~\citep{ho2020denoising}; \emph{STASY}~\citep{kim2022stasy} uses a score-based diffusion model~\citep{song2019generative}; \emph{TabSyn} leverages a diffusion model within a Transformer-based variational autoencoder~\citep{zhang2023mixed}; \emph{TabPFN}~\cite{tabpfn} is a Transformer-based tabular foundation model pretrained on synthetic datasets generated from a wide range of causal graphs; \emph{Forest-VP} and \emph{Forest-Flow}~\citep{jolicoeur2023generating} use tree-based~\citep{chen2015xgboost} deep diffusion and flow matching models~\citep{lipman2022flow}; \emph{CTGAN} utilizes a conditional generative adversarial network~\citep{xu2019modeling}; \emph{TVAE} utilizes a variational autoencoder~\citep{xu2019modeling}. Additionally, \emph{GOGGLE}~\citep{liu2022goggle} uses an encoder-decoder model to generate tabular data, where the decoder is a graph neural network to capture the dependencies between features. 
Recent work on causal normalizing flows~\citep{javaloy2024causal} also advances the modeling of complex causal relationships in generative settings.
Given the increasing number of tabular synthesis models, we mainly mentioned the representative ones for each category and benchmarked most of them in the experiments. However, we highly recommend survey papers~\citep{borisov2022deep,li2023graph,fang2024large} for more details on tabular synthesis models.
\section{High-Order Structural Causal Information}
\label{sec:preliminaries}
A fundamental problem in modeling tabular data is the lack of prior knowledge about their structures and high-order information~\citep{borisov2022deep,fang2024large}. Natural and common prior knowledge in tabular domain can be causal dependencies in forms of causal graphs~\citep{peters2017elements,glymour2019review}. Real-world data are generated by certain underlying mechanisms, which can be qualitatively described using causal graphs. As for tabular data whose columns are variables of interests, \emph{causal graphs} are directed graphs where the nodes are variables and the directed edges represent causal relationships between the columns. Different from mere pair-wise information (e.g., correlations), such causal relationships represent high-order structural information. Capturing this type of high-order information requires methods that go beyond pair-wise reasoning. In our framework, we assume the causal graphs are directed acyclic graphs (DAGs) without unknown confounders. These are common assumptions in causal machine learning~\citep{peters2017elements,scholkopf2021toward} under which the studies are significantly supported by well-studied properties and theories as well as reliable methods and considerable applications. 
Given a causal DAG, the causal information is the high-order statistical information which under proper assumptions has asymmetric properties implying direct causes and effects in the data generation process. 
We categorize causal information into three hierarchical levels: causal skeleton, Markov equivalence class, and causal DAG, each capturing progressively richer structural details. While lower levels are easier to obtain, they convey less causal information than the higher ones. These levels correspond to different forms of causal graphs, as illustrated in Figure~\ref{fig:three_level}.
Throughout, ``complexity'' is used in a structural sense: the number of nodes, edge count and density, and the induced multivariate d-separation and direction relations. A modest node count can still exhibit substantial structural complexity through these factors.
\begin{figure*}
%\vspace{-5pt}
\begin{subfigure}{.32\textwidth}
  \centering
    \begin{tikzpicture}[
			scale=0.4,
            > = stealth, % arrow head style
            shorten > = 1pt, % don't touch arrow head to node
            auto,
            node distance = 3cm, % distance between nodes
            semithick % line style
        ]

        \tikzstyle{every state}=[
            draw = black,
            thick,
            minimum size = 8mm
        ]

        \node[state] (X) at  (-3,3){$X$};
        \node[state] (Y) at  (3,3) {$Y$};
        \node[state] (Z) at  (-0,3){$Z$};
        \node[state] (W) at  (-0,0) {$W$};

        \path[-] (X) edge node {} (W);
        \path[-] (Y) edge node {} (W);
        \path[-] (X) edge node {} (Z);
        \path[-] (Z) edge node {} (Y);
        
    \end{tikzpicture}
  \caption{Causal skeleton level: Undirected causal graphs show the connectivity entailing whether pairs of nodes are d-separated or not.}
  \label{fig:cskeleton}
\end{subfigure}
\hfill
\begin{subfigure}{.32\textwidth}
	\centering
	\begin{tikzpicture}[
			scale=0.4,
            > = stealth, % arrow head style
            shorten > = 1pt, % don't touch arrow head to node
            auto,
            node distance = 3cm, % distance between nodes
            semithick % line style
        ]

        \tikzstyle{every state}=[
            draw = black,
            thick,
            minimum size = 8mm
        ]

        \node[state] (X) at  (-3,3){$X$};
        \node[state] (Y) at  (3,3) {$Y$};
        \node[state] (Z) at  (-0,3){$Z$};
        \node[state] (W) at  (-0,0) {$W$};

        \path[->] (X) edge node {} (W);
        \path[->] (Y) edge node {} (W);
        \path[-] (X) edge node {} (Z);
        \path[-] (Z) edge node {} (Y);
        
    \end{tikzpicture}  
    \caption{Markov equivalent class level: Completed partially oriented directed acyclic graphs provide 
    % show the connectivity and some causal directions with the 
    all d-separation information.} 
    \label{fig:cpdag}
\end{subfigure}
\hfill
\begin{subfigure}{.3\textwidth}
  \centering
  \begin{tikzpicture}[
			scale=0.4,
            > = stealth, % arrow head style
            shorten > = 1pt, % don't touch arrow head to node
            auto,
            node distance = 3cm, % distance between nodes
            semithick % line style
        ]

        \tikzstyle{every state}=[
            draw = black,
            thick,
            minimum size = 8mm
        ]

        \node[state] (X) at  (-3,3){$X$};
        \node[state] (Y) at  (3,3) {$Y$};
        \node[state] (Z) at  (-0,3){$Z$};
        \node[state] (W) at  (-0,0) {$W$};

        \path[->] (X) edge node {} (W);
        \path[->] (Y) edge node {} (W);
        \path[<-] (X) edge node {} (Z);
        \path[->] (Z) edge node {} (Y);
        
    \end{tikzpicture}
 \caption{Causal DAG (directed acyclic graph) level: It provides all d-separation and causal-direction information.}
  \label{fig:cdag}
\end{subfigure}%
\vspace{-3pt}
\caption{Three levels of high-order structural causal information.}\label{fig:three_level}
\vspace{-10pt}
\end{figure*}

{\bf Level 1: Causal skeleton.}
\emph{Causal skeletons} are the undirected graphs of causal DAGs~\citep{spirtes2000causation}, and describe the connectivity of the nodes. At this level, causal information represents \emph{d-separation} relationships as a type of high-order information. Given a causal DAG, two nodes $X$ and $Y$ are d-separated by a set $\boldsymbol{Z}$ if and only if for each path between $X$ and $Y$, there is a chain $\cdot \rightarrow Z \rightarrow  \cdot$ or a fork $\cdot \leftarrow Z \rightarrow  \cdot$ such that $Z$ is in $\boldsymbol{Z}$; or it contains a collider $\cdot \rightarrow W \leftarrow  \cdot$  such that $W$ and its descendants are not in $\boldsymbol{Z}$~\citep{peters2017elements}. Therefore, given a causal skeleton, we know the nodes that are d-separated; and further under the \emph{causal sufficiency assumption}, i.e., there are no unmeasured confounders (the same common parent of children nodes), the connectivity of two nodes in causal skeletons infers whether a pair of variables are causally dependent or not. 

{\bf Level 2: Markov equivalent class.} At this level, we not only know the d-separation relationships of node pairs, but we also know by which nodes set they are d-separated. Two DAGs are (Markov) equivalent if and only if they have the same d-separation relationships. The Markov equivalent class can be uniquely represented by a completed partially directed acyclic graph (CPDAG) under proper assumptions~\citep{meek1995causal,andersson1997characterization}, which can provide complete d-separation information alongside certain causal directions, enriching the information of the causal skeleton.

{\bf Level 3: Causal DAG.} At this level, all information about connectivity and causal directions is summarized in causal DAGs. This causal information goes beyond knowing all the d-separation relationships. Given the nodes in a causal DAG, we know all of their asymmetric causal relationships, i.e., their direct causes and direct effects.

\paragraph{Applications enabled by the framework.}
Beyond benchmarking for its own sake, the framework enables:
(1) structure-aware model selection when preserving causal dependencies matters;
(2) diagnosis of why a generator fails e.g. strong low-order statistics but poor d-separation or directionality;
(3) development of causal-aware tabular generators and training objectives; and
(4) safer use of synthetic data for downstream decision-support tasks such as interventional and counterfactual estimation.

\section{High-Order Structural Causal Benchmark Framework}
\label{sec:benchmark}
To support the development of reliable tabular synthesis models, especially those based on deep generative frameworks, a rigorous evaluation is essential. In particular, intrinsic evaluation, assessing the quality of synthetic data independently of specific downstream tasks, plays a critical role~\citep{van2023beyond}. As discussed in Section~\ref{sec:preliminaries}, high-order structural causal information provides a principled basis for intrinsic evaluation by directly measuring a model’s ability to capture complex causal dependencies~\citep{scholkopf2021toward}.

To enable such evaluation, our benchmark framework (Figure~\ref{fig:framework}) generates synthetic benchmark datasets using predefined causal graphs and diverse data generation processes (Section~\ref{sec:benchmark_data}). These causal graphs provide the ground-truth structural labels required for metric computation. We then apply causal discovery methods to extract different levels of causal information\footnote[1]{Causal skeletons and Markov equivalence classes can be identified by constraint-based methods~\citep{spirtes2000causation}, while causal directions in DAGs require functional causal model-based methods~\citep{glymour2019review}. See Appendix~\ref{app:causal_discovery} for an overview of causal discovery and identifiability assumptions.} from both benchmark and synthetic datasets, and we compare each of them to the ground truth labels, allowing us to define high-order evaluation metrics at multiple levels (Section~\ref{sec:benchmark_task}). We also include causal inference tasks as complementary downstream evaluations. For clarity, we use the term \emph{benchmark datasets} to refer to synthetic data generated from causal graphs for training synthesis models, and \emph{synthetic datasets} to refer to the outputs of these trained models.

\subsection{Generation of benchmark datasets} \label{sec:benchmark_data}
The validity of benchmark datasets requires that (i) data are generated according to causal directed acyclic graphs and (ii) causal information (causal skeletons or directions) is identifiable from the data by causal discovery methods under proper assumptions.
Condition (i) requires that each benchmark dataset has a corresponding causal DAG and a causal DAG can be used to generate a dataset for evaluation. Regarding Condition (i), the data generation process according to a causal DAG is
\begin{eqnarray}
    X_i = f_i(X_i^{\texttt{prt}}, E_{X_i}) \label{eqn:generation},
\end{eqnarray}
where $X_i$ is a variable in the tabular dataset and a node in a causal DAG, $X_i^{\texttt{prt}}$ is the parents of $X_i$ in the graph, $f_i$ is the causal functional relationship between parent and child variables, and $E_{X_i}$ is noise that is independent of $X_i^{\texttt{prt}}$. 
Condition (ii) requires that causal discovery methods validly recover causal information from a benchmark dataset. 
We limit data generation processes, i.e., the functional relationships and noise distributions in~\eqref{eqn:generation}, to the identifiable ones for causal discovery methods under proper assumptions~\citep{glymour2019review}. 
Therefore, we categorize the data generation processes by their functional relationships and noise distributions. 
More specifically, our data generation processes use three types of functional relationships -- linear (L), sigmoid (S), and neural network-based (N) -- denoted by $f_i$, combined with two types of noise distributions for $E_{X_i}$: Gaussian (G) and Uniform (U). 
For example, we denote the benchmark dataset generated with a linear functional relationship and Gaussian variables by ``LG''. We modify~\href{https://fentechsolutions.github.io/CausalDiscoveryToolbox/html/index.html}{\texttt{CausalDiscoveryToolbox}} to randomly generate benchmark datasets with continuous values according to a given causal DAG. 
DAGs are sampled with a per-node in-degree bound of 2. Across the 10 generated graphs, edge counts are $9.0 \pm 2.8$ for 10-node DAGs (range 5--14) and $18.5 \pm 2.8$ for 20-node DAGs (range 15--23).

The causal mechanisms for linear, sigmoid, and neural network-based functional relationships are 
\begin{eqnarray}
    X_i &=& W_i \cdot X_i^{\texttt{prt}} + E_{X_i}; \\
    X_i &=& W_i \cdot \sigma(X_i^{\texttt{prt}}) + E_{X_i}; \\
    X_i &=& W_{1i} \cdot  \sigma(W_{2i} \cdot  (X_i^{\texttt{prt}}\oplus E_{X_i})); 
\end{eqnarray}
where $\oplus$ is concatenation, and $W_i$, $W_{1i}$, and $W_{2i}$ are weight matrices.

To generate a discrete value with $K$ categories of $X_i^{\texttt{disc.}}$, we first generate a continuous value, then compute the probability of each category as~\eqref{eqn:prob_k}, and sample from the categorical distribution as~\eqref{eqn:onehot},
\begin{eqnarray}
    \texttt{prob}_k &\coloneqq& \texttt{softmax}(\sigma(W_{i,k} \cdot X_i)); \label{eqn:prob_k}\\
    X_i^{\texttt{disc.}}&\sim&\mathcal{C}(\texttt{prob}_1,...,\texttt{prob}_K),\label{eqn:onehot}
\end{eqnarray}
where $X_i$ is a continuous-valued variable, $X_i^{\texttt{disc.}}$ is a discrete-valued variable, $\sigma$ is the sigmoid function, $W_{i,k}$ is a random weight for each category of $X_i^{\texttt{disc.}}$, $\mathcal{C}$ denotes a categorical distribution with parameters $\texttt{prob}_k$ for $k=1,...,K$. The results of applying this procedure for the linear gaussian setting are available in Table \ref{tab:discrete}.

\subsection{Benchmark metrics}\label{sec:benchmark_task}
Benchmark datasets are generated from predefined DAGs and used to train tabular synthesis models, and their corresponding causal DAGs are used to derive ground-truth labels of causal information for evaluation. We define high-order structural causal metrics by applying different causal discovery methods to infer causal information (typically unavailable) at different levels from synthetic data. 
Roughly speaking, our metrics compare how well causal information can be recovered by measuring the deviation of discovered DAGs - from both benchmark and synthetic data -  from the ground-truth causal labels; cf.~Figure~\ref{fig:framework}. 
Besides causal information at different levels, metrics also indicate model capabilities to capture joint or individual information, depending on the task. For example, individual causal information can be d-separations and causal directions. Joint causal information is based on the aggregation and integration of individual causal information.

{\bf Metrics on causal skeletons.}
Causal skeletons can be determined by constraint-based causal discovery methods. In our experiments, we apply PC algorithm~\citep{spirtes2000causation} to benchmark datasets and synthetic datasets and then get the adjacency matrices of causal skeletons. Furthermore, given the resulting adjacency matrices and the adjacency matrices derived from ground-truth causal DAGs, structural Hamming distance (SHD), recall, precision, and F1 score can be used to measure the differences between the resulted and the ground-truth adjacency matrices. Such metrics also indicate model capability of capturing joint causal information, because causal skeletons are constructed by summarizing multiple d-separations.

{\bf Metrics on conditional independence relationships.} \label{sec:metric_ci}
Under causal sufficiency, faithfulness, and causal Markov assumptions, conditional independence in data implies d-separation in a causal graph~\citep{spirtes2000causation}. 
We use conditional independence relationships for benchmarking on the Markov equivalent level.
The task based on individual causal information without requiring integrating d-separations. We first select a d-separation and d-connection set with the same set sizes denoted by $\mathbf{D} = \{(X_i,Y_i,\mathbf{S}_i)\}_{i=1:N}$, where $X_i$ and $Y_i$ are either d-connected or d-separated conditioning on the set $\mathbf{S}_i$. 
We then apply conditional independence tests to the selected subsets of benchmark and synthetic datasets and get results $\mathbf{C}^{\texttt{ref}} = \{c^{\texttt{ref}}_i: 0 \text{ or } 1\}_{i\in\mathbf{D}}$ and $\mathbf{C}^{\texttt{syn}} = \{c^{\texttt{syn}}_i: 0 \text{ or } 1\}_{i\in\mathbf{D}}$, where $0$ and $1$ represent conditional dependence and independence respectively; and derive the ground-truth conditional independence relationships from ground-truth causal DAGs denoted by $\mathbf{C}^{\texttt{gt}} = \{c^{\texttt{gt}}_i: 0 \text{ or } 1\}_{i\in\mathbf{D}}$. Considering the evaluation on this level as the evaluation of a binary classification problem, Area Under the Curve (AUC) scores of Receiver Operating Characteristic (ROC) curves are used as a metric. Specifically, we are comparing the discovered conditional dependence and independence labels $\mathbf{C}^{\texttt{ref}}$ and $\mathbf{C}^{\texttt{syn}}$ against the ground truth labels $\mathbf{C}^{\texttt{gt}}$. 

{\bf Metrics on causal directions.}
As for methods identifying causal directions, bivariate causal discovery methods~\citep{hoyer2008nonlinear,janzing2012information} are commonly available. Different from the other metrics, the metric using bivariate causal discovery methods is based on the bivariate setting. We first select a set of edges from the ground-truth causal DAGs denoted by $\mathbf{E} = \{(X_i,Y_i)\}_{i=i:N}$, of which $X_i$ and $Y_i$ are d-separated after removing the edges between them. In this way, we can apply bivariate causal discovery methods to the data of $X_i$ and $Y_i$ without the impact of the other paths between them on the causal direction of the edge between them. We apply bivariate causal discovery methods on the selected subsets of benchmark and synthetic datasets and get the results denoted by $\mathbf{E}^{\texttt{ref}} = \{e^{\texttt{ref}}_i\!: 0 \text{ or } 1\}_{i\in\mathbf{E}}$ and $\mathbf{E}^{\texttt{syn}} = \{e^{\texttt{syn}}_i\!: 0 \text{ or } 1\}_{i\in\mathbf{E}}$; and derive the ground-truth conditional independence relationships from ground-truth causal DAGs denoted by $\mathbf{E}^{\texttt{gt}} = \{e^{\texttt{gt}}_i\!: 0 \text{ or } 1\}_{i\in\mathbf{E}}$, where $0$ and $1$ represent different causal directions. The metric on this level is the accuracy of the predicted results, $\mathbf{E}^{\texttt{ref}}$ and $\mathbf{E}^{\texttt{syn}}$, compared to the ground-truth labels $\mathbf{E}^{\texttt{gt}}$. 
As a result, the evaluation with bivariate causal discovery methods is based on individual causal information. 
In addition to bivariate methods, LiNGAM (Linear Non-Gaussian Acyclic Model)-based approaches~\citep{shimizu2006linear,shimizu2011directlingam} can identify causal directions in the multivariate linear non-Gaussian setting. In this case, SHD, precision, recall, and F1 score are calculated by comparing the resulting fully oriented causal graphs to the ground truth DAGs, similar to the metrics used at the causal skeleton level. This evaluation reflects joint causal information.

{\bf Metrics on downstream tasks.}
Evaluating tabular synthesis models on downstream tasks helps assess how well the generated data supports causal reasoning and decision-making. In our framework, this involves training Structural Causal Models (SCMs) on synthetic data and evaluating their performance on downstream causal inference tasks using held-out benchmark data~\citep{zhang2023mixed,fang2024large}. 
We focus on interventional and counterfactual inference tasks, as their performance directly reflects a model’s ability to capture essential causal information, which is crucial for our benchmarking objective. 
Our evaluation and metrics are inspired by~\cite{chen2023structured}. Firstly, benchmark and synthetic data are used for training SCMs given corresponding causal graphs. In the interventional inference task, we perform a series of interventions on each variable in the causal graph, one at a time, and utilize the trained SCM to compute the resulting interventional distributions over the remaining variables. Furthermore, we compute the average differences between the expectation of interventional distributions generated by SCM models trained on synthetic data and benchmark data. For the counterfactual inference task, we generate new observations with the ground-truth SCM for each causal graph. We then compute their counterfactual values with trained SCMs by imposing interventions on each variable individually. The metric is based on the average differences of average counterfactual values between SCM models trained on synthetic and benchmark data. These metrics are detailed in Section~\ref{sec:exp_setting}.

\section{Experiments}\label{sec:experiments}
We begin by outlining the evaluation procedures and experimental settings. 
Section~\ref{sec:exp_results} presents benchmarking results of state-of-the-art tabular synthesis models on high-order structural causal tasks using synthetic datasets generated from known causal DAGs. We further evaluate these methods on downstream causal inference tasks and examine their performance on a widely adopted real-world dataset, highlighting the capabilities and limitations of each baseline method. 
Additional experimental details, benchmark configurations, implementation specifics, and supplementary results (including metrics such as $\alpha$-precision, $\beta$-recall, single-variable density estimation, and pairwise correlation scores~\citep{alaa2022faithful,zhang2023mixed}) are provided in Appendix~\ref{app:exp_details}, alongside further insights into the high-order metrics introduced in Section~\ref{sec:benchmark}.

\subsection{Experimental settings}\label{sec:exp_setting}
Our baseline methods cover Transformer-based, DFM-based, Generative Adversarial and Variational Autoencoder methods which are TabSyn~\citep{zhang2023mixed}, STASY~\citep{kim2022stasy}, TabDDPM~\citep{kotelnikov2023tabddpm}, CoDi~\citep{lee2023codi}, GReaT~\citep{borisov2022deep}, CTGAN~\citep{xu2019modeling}, TVAE~\citep{xu2019modeling} and TabPFN \citep{tabpfn}. 
Our baseline set is chosen to be representative of the major architecture families currently used in tabular synthesis i.e. diffusion / score-based, Transformer-based, and GAN / VAE together with the prior-fitted tabular foundation model TabPFN. The benchmark itself is model-agnostic and supports straightforward extension to newer tabular-synthesis models.
Firstly, to benchmark baseline methods on high-order structural causal information, 
$N_g$ causal DAGs 
$\mathcal{G}^{\texttt{gt}} = \{G^{\texttt{gt}}_g\}_{g=1:N_g}$
are randomly generated and each causal DAG is used for generating benchmark datasets 
$\mathcal{D}_g^{\texttt{gt}} = \{D^{\texttt{gt}}_{g,m}\}_{m \in \omega}$ 
with different causal mechanisms 
$\omega=\{\text{LG, LU, SG, NG}\}$. 
We then train baseline methods on benchmark datasets $D^{\texttt{gt}}_{g,m}$ and generate synthetic datasets $\mathcal{D}_g^{\texttt{syn}} = \{D^{\texttt{syn}}_{g,m}\}_{m\in\omega}$ for each $G^{\texttt{gt}}_g$. 
Secondly, with the causal DAGs, benchmark datasets, and synthetic datasets, causal information is identified by causal discovery methods. And causal information, such as adjacency matrices, conditional independence relationships, and predicted edge directions, is denoted by 
$Q^{\texttt{ref}}_{g,m} \coloneqq \texttt{CD}(D^{\texttt{gt}}_{g,m}); ~~~Q^{\texttt{syn}}_{g,m}\coloneqq \texttt{CD}(D^{\texttt{syn}}_{g,m}), $
where $\texttt{CD}$ are causal discovery methods, and ground-truth causal information is derived from causal DAGs, denoted by $Q^{\texttt{gt}}_g$. We include the results on benchmark datasets as a reference ceiling to contextualize performance on synthetic data. In all cases or causal levels, causal information discovered from both benchmark and synthetic data is evaluated against ground-truth DAGs, we do not compare one discovered DAG to another.
Formally, for each metric $M_i \in \mathcal{M}= \{\text{F1, SHD, others in Section~\ref{sec:benchmark_task}}\}$, we compute
\sloppy$R_{g,m}^{\texttt{ref}} = M_i(Q^{\texttt{ref}}_{g,m},Q^{\texttt{gt}}_{g});~~~R_{g,m}^{\texttt{syn}} = M_i(Q^{\texttt{syn}}_{g,m},Q^{\texttt{gt}}_{g}),$ to evaluate all baseline methods.
For different evaluation purposes, we can aggregate the metric values along different indices and make conclusions. In our experiments, we compute average metric values over all causal DAGs for each causal mechanism:
\begin{eqnarray*}
R_{m}^{\texttt{ref}} = \texttt{AVE}_{g}(R_{g,m}^{\texttt{ref}})
\quad \text{and} \quad
R_{m}^{\texttt{syn}} = \texttt{AVE}_{g}(R_{g,m}^{\texttt{syn}}).
\end{eqnarray*}

{\bf Training and evaluation sample sizes.}
Models are trained on the full 17{,}117-row benchmark datasets; causal metrics are computed on 15{,}000-row evaluation subsets, with 10 bootstrap samples at the skeleton level.
{\bf Benchmark on the level of causal skeleton.} 
For each causal mechanism, we generate $10$ benchmark datasets (according to 10 causal DAGs) with N variables, a flexible parameter in our framework, and around 17,000 samples. For computing metric values, we get 10 bootstrapping datasets with sample size 15,000 for each benchmark and synthetic dataset, and apply PC algorithm to obtain causal information quantities, denoted by $Q^{\texttt{ref}}_{g,m,b}$ and $Q^{\texttt{syn}}_{g,m,b}$ where the bootstrapping index is $b = \{1,...,10\}$. The metric value on each causal DAG is the average SHD, recall, precision and F1 scores over $10$ bootstrapping datasets, $R_{g,m}^{\texttt{ref}} = \texttt{AVE}_{b=1:10}(R_{g,m,b}^{\texttt{ref}});R_{g,m}^{\texttt{syn}} = \texttt{AVE}_{b=1:10}(R_{g,m,b}^{\texttt{syn}})$. 
The average and standard deviation of metric values reported in Table~\ref{tab:combined_lg_lu_final} are computed based on 10 benchmark datasets with 10 continuous variables for each causal mechanism. Additional experiment results with more continuous variables available in appendix~\ref{app:exp_details}.

{\bf Benchmark on the level of Markov equivalent class.}
To evaluate causal information on the Markov equivalent class level, we find all d-separations with minimal conditional sets between each pair of nodes in a causal graph and then apply conditional independence tests to the corresponding sets on synthetic datasets. 
We use the same procedures as the experiments on causal skeleton level to train baseline methods and generate synthetic datasets. 
Since the experimental results show minimal variation across different bootstrapped datasets, we omit bootstrapping at this level and use 15,000 samples for the tests.

{\bf Benchmark on the level of causal direction.}
To evaluate causal directions using only data from a pair of variables, we select edges between nodes that become d-separated once the edge connecting them is removed; this ensures that other paths in the graph do not confound the pairwise causal relationship.
We employ three commonly used training-free bivariate causal discovery methods RECI~\citep{blobaum2018cause}, IGCI~\citep{janzing2012information}, and CDS~\citep{fonollosa2019conditional}. We report the best of the three for each dataset/mechanism as the pairwise direction score. The aggregate is intended to reduce dependence on any single bivariate method rather than to claim one method dominates universally; per-method results are reported in Appendix~\ref{app:exp_details}.
Separately, we use LiNGAM~\citep{shimizu2006linear} as an additional multivariate direction evaluation. Each evaluation is conducted using 15{,}000 samples.

{\bf Benchmark on downstream tasks.} 
Our evaluation is largely inspired by~\cite{chen2023structured}. As mentioned in Section~\ref{sec:benchmark_task}, same SCM models are trained on benchmark data and synthetic data together given the underlying causal DAGs. In the interventional inference task, for each causal graph, we take 10 interventions on each variable and approximate the estimated interventional distributions with 1,000 samples. In the counterfactual inference task, for each causal graph, we impose 10 interventions on each variable and generate 1,000 benchmark data as the new observations for computing counterfactual values. Next, we compare the results on interventional and counterfactual tasks subject to the same interventions. Considering the results of the models trained on benchmark data as the ground-truth, the metric is the average mean absolute errors (AMAE) over all variables, 
\begin{equation}
\begin{aligned}
\text{AMAE-\texttt{syn}} &= \frac{1}{V}\sum_{v\in \boldsymbol{V}} \text{MAE-\texttt{syn}}(v),\\
\text{MAE-\texttt{syn}}(v) &=
\frac{1}{(V\!-\!1)K N}
\!\sum_{i\in \boldsymbol{V}\setminus \{v\}}
\sum_{d\in \boldsymbol{S}_r}
\left|
\sum_{s=0}^{N-1} x^{\texttt{ref}}_{s,d,i}
-\sum_{s=0}^{N-1} x^{\texttt{syn}}_{s,d,i}
\right|,
\end{aligned}
\end{equation}
where $x^{\texttt{ref}}_{s,d,i}$ is the reference ground-truth result and $x^{\texttt{syn}}_{s,d,i}$ is the result of the models trained on synthetic data;
$s$ denotes different samples;
$v$ denotes that the interventions are imposed on variable $v$ chosen from the variable set $\boldsymbol{V}$ with $V$ variables;
$d$ denotes the intervention value that is taken from a set $\boldsymbol{S}_r$ with sample size $K$; and 
$i$ denotes the variable of which the interventional distribution or counterfactual value is computed for evaluation. We compute the mean and standard deviation of the average differences with 10 random seeds. 

\begin{table*}[ht]
    \centering
    \caption{Benchmark results under linear Gaussian (LG), linear uniform (LU), sigmoid Gaussian (SG) and neural network Gaussian causal mechanisms. The LU case is also visualized in Figure~\ref{fig:lu_radar}. Complete results are detailed in Appendix~\ref{app:exp_details}.}
    \vspace{-5pt}
    \label{tab:combined_lg_lu_final}
    \setlength{\tabcolsep}{3pt}
    \renewcommand{\arraystretch}{0.7}
    \resizebox{\linewidth}{!}{
    \begin{tabular}{c|l|cc|c|c|ccc|c|c}
    \toprule
    & \multirow{2}{*}{Model} & \multicolumn{2}{c|}{Low-order} & Skeleton & MEC & \multicolumn{3}{c|}{Causal direction level} & Intervention & Counterfact \\
    & & Col.\ ER ($\downarrow$) & Pair.\ ER ($\downarrow$) & F1 ($\uparrow$) & AUC ($\uparrow$) & SHD ($\downarrow$) & ACC ($\uparrow$) & F1 ($\uparrow$) & AMAE ($\downarrow$) & AMAE ($\downarrow$) \\
    \midrule
    \multirow{8}{*}{\rotatebox{90}{LG: linear Gaussian}} 
& ref.     & $0.00 \pm 0.00$ & $0.00 \pm 0.00$ & $0.90 \pm 0.06$ & $0.972$ & \cellcolor{gray!20}$15.42 \pm 7.01$ & \cellcolor{gray!20}$0.500$ & \cellcolor{gray!20}$0.38 \pm 0.06$ & $3.16 \pm 0.2$ & $0.04 \pm 0.0$ \\
\cmidrule(r){2-11}
& TabSyn   & $2.11 \pm 1.08$ & $0.61 \pm 0.21$ & $0.71 \pm 0.12$ & $0.927$ & \cellcolor{gray!20}$27.74 \pm 5.14$ & \cellcolor{gray!20}$0.500$ & \cellcolor{gray!20}$0.26 \pm 0.07$ & $4.73 \pm 1.8$ & $0.56 \pm 0.4$ \\
& STASY    & $12.42 \pm 3.27$ & $1.31 \pm 0.81$ & $0.68 \pm 0.17$ & $0.930$ & \cellcolor{gray!20}$31.92 \pm 4.55$ & \cellcolor{gray!20}$\tablebf{0.696}$ & \cellcolor{gray!20}$0.21 \pm 0.07$ & $26.66 \pm 7.1$ & $0.65 \pm 0.4$ \\
& TabDDPM  & $\tablebf{0.69 \pm 0.12}$ & $0.62 \pm 0.54$ & $0.70 \pm 0.12$ & $0.814$ & \cellcolor{gray!20}$26.64 \pm 10.34$ & \cellcolor{gray!20}$0.536$ & \cellcolor{gray!20}$0.24 \pm 0.09$ & $4.13 \pm 1.2$ & $1.12 \pm 1.3$ \\
& CoDi     & $4.42 \pm 0.83$ & $0.80 \pm 0.43$ & $0.72 \pm 0.10$ & $0.917$ & \cellcolor{gray!20}$29.66 \pm 5.12$ & \cellcolor{gray!20}$0.589$ & \cellcolor{gray!20}$0.24 \pm 0.08$ & $5.25 \pm 1.6$ & $0.67 \pm 0.8$ \\
& GReaT    & $8.41 \pm 0.73$ & $0.76 \pm 0.26$ & $0.75 \pm 0.04$ & $0.921$ & \cellcolor{gray!20}$18.18 \pm 8.97$ & \cellcolor{gray!20}$0.554$ & \cellcolor{gray!20}$0.36 \pm 0.06$ & $9.77 \pm 0.7$ & $0.93 \pm 0.5$ \\
& CTGAN    & $4.70 \pm 0.78$ & $3.76 \pm 0.63$ & $0.46 \pm 0.06$ & $0.566$ & \cellcolor{gray!20}$36.00 \pm 6.40$ & \cellcolor{gray!20}$0.554$ & \cellcolor{gray!20}$0.20 \pm 0.06$ & $15.02 \pm 8.9$ & $8.58 \pm 7.7$ \\
& TVAE     & $4.51 \pm 1.64$ & $1.93 \pm 0.64$ & $0.58 \pm 0.06$ & $0.703$ & \cellcolor{gray!20}$29.60 \pm 8.47$ & \cellcolor{gray!20}$0.536$ & \cellcolor{gray!20}$0.20 \pm 0.05$ & $9.52 \pm 5.2$ & $5.22 \pm 3.9$ \\
& TabPFN & $1.63 \pm 0.17$ & $\tablebf{0.30 \pm 0.09} $ &   $\tablebf{0.88 \pm 0.07}$ &  $\tablebf{0.970}$ & \cellcolor{gray!20}$\tablebf{14.66 \pm 6.16} $ &   \cellcolor{gray!20}$0.375 $ &  \cellcolor{gray!20}$\tablebf{0.30 \pm 0.09}$ &  $\tablebf{3.35 \pm 0.3 }$ & $ \tablebf{0.21 \pm 0.1 }  $\\
\midrule
\midrule
\multirow{8}{*}{\rotatebox{90}{LU: linear uniform}} 
    & ref.     & $0.00 \pm 0.00$ & $0.00 \pm 0.00$ & $0.89 \pm 0.07$ & $0.967$ & $1.04 \pm 0.85$ & $1.000$ & $0.94 \pm 0.04$ & $3.07 \pm 0.2$ & $0.03 \pm 0.0$ \\
    \cmidrule(r){2-11}
    & TabSyn   & $1.88 \pm 0.87$ & $0.45 \pm 0.28$ & $0.71 \pm 0.08$ & $0.871$ & $21.66 \pm 5.75$ & $0.946$ & $0.41 \pm 0.06$ & $4.21 \pm 1.1$ & $0.45 \pm 0.7$ \\
    & STASY    & $11.90 \pm 5.72$ & $1.18 \pm 0.59$ & $0.67 \pm 0.10$ & $\tablebf{0.936}$ & $20.66 \pm 4.99$ & $0.946$ & $0.45 \pm 0.10$ & $23.86 \pm 11.1$ & $0.44 \pm 0.3$ \\
    & TabDDPM  & $\tablebf{0.98 \pm 0.63}$ & $1.07 \pm 2.27$ & $0.77 \pm 0.07$ & $0.815$ & $15.86 \pm 8.74$ & $\tablebf{1.000}$ & $0.51 \pm 0.14$ & $\tablebf{3.48 \pm 0.6}$ & $0.46 \pm 0.7$ \\
    & CoDi     & $8.01 \pm 1.55$ & $1.75 \pm 1.52$ & $0.74 \pm 0.09$ & $0.902$ & $18.54 \pm 5.08$ & $0.911$ & $0.45 \pm 0.10$ & $3.82 \pm 0.6$ & $0.58 \pm 0.4$ \\
    & GReaT    & $9.56 \pm 0.73$ & $0.56 \pm 0.22$ & $0.70 \pm 0.05$ & $0.860$ & $13.62 \pm 9.24$ & $0.929$ & $0.57 \pm 0.11$ & $11.20 \pm 1.2$ & $0.58 \pm 0.4$ \\
    & CTGAN    & $4.71 \pm 0.50$ & $3.52 \pm 0.50$ & $0.51 \pm 0.08$ & $0.588$ & $33.64 \pm 5.02$ & $0.821$ & $0.26 \pm 0.07$ & $10.89 \pm 3.2$ & $4.96 \pm 3.3$ \\
    & TVAE     & $6.33 \pm 1.69$ & $2.52 \pm 1.17$ & $0.55 \pm 0.07$ & $0.669$ & $26.62 \pm 8.15$ & $0.839$ & $0.29 \pm 0.08$ & $12.22 \pm 3.4$ & $5.40 \pm 3.6$ \\
    & TabPFN & $2.04 \pm 0.40  $ & $\tablebf{0.30 \pm 0.10 }$ &   $\tablebf{0.88 \pm 0.06}$ &  $0.928$ & $\tablebf{2.44 \pm 2.62} $ &   $0.964 $ &  $\tablebf{0.88 \pm 0.11}$ &  $3.87 \pm 0.5 $ & $ \tablebf{0.18 \pm 0.1 }  $\\

\midrule
\midrule
\multirow{8}{*}{\rotatebox{90}{SG: sigmoid Gaussian}} 
& ref.     & $0.00 \pm 0.00$ & $0.00 \pm 0.00$ & $0.95 \pm 0.03$ & $0.982$ & \cellcolor{gray!20}$25.02 \pm 11.73$ & \cellcolor{gray!20}$1.000$ & \cellcolor{gray!20}$0.19 \pm 0.06$ & $3.3 \pm 0.1$ & $0.32 \pm 0.2$ \\
\cmidrule(r){2-11}
& TabSyn   & $1.83 \pm 0.84$ & $0.49 \pm 0.20$ & $0.92 \pm 0.02$ & $0.974$ & \cellcolor{gray!20}$28.44 \pm 10.14$ & \cellcolor{gray!20}$\tablebf{1.000}$ & \cellcolor{gray!20}$0.16 \pm 0.06$ & $4.6 \pm 1.1$ & $0.88 \pm 0.4$ \\
& STASY    & $12.48 \pm 5.13$ & $1.52 \pm 0.43$ & $0.88 \pm 0.07$ & $0.963$ & \cellcolor{gray!20}$32.58 \pm 6.72$ & \cellcolor{gray!20}$0.982$ & \cellcolor{gray!20}$0.18 \pm 0.06$ & $25.7 \pm 10.5$ & $1.46 \pm 0.8$ \\
& TabDDPM  & $\tablebf{0.85 \pm 0.14}$ & $0.40 \pm 0.30$ & $0.92 \pm 0.03$ & $\tablebf{0.982}$ & \cellcolor{gray!20}$28.68 \pm 10.54$ & \cellcolor{gray!20}$0.893$ & \cellcolor{gray!20}$0.17 \pm 0.05$ & $4.2 \pm 0.4$ & $1.20 \pm 0.6$ \\
& CoDi     & $5.96 \pm 2.53$ & $1.14 \pm 0.29$ & $0.87 \pm 0.06$ & $0.956$ & \cellcolor{gray!20}$31.00 \pm 9.69$ & \cellcolor{gray!20}$0.964$ & \cellcolor{gray!20}$0.16 \pm 0.08$ & $10.2 \pm 4.2$ & $0.94 \pm 0.4$ \\
& GReaT    & $8.51 \pm 2.03$ & $1.45 \pm 0.72$ & $0.87 \pm 0.07$ & $0.964$ & \cellcolor{gray!20}$26.08 \pm 12.33$ & \cellcolor{gray!20}$0.946$ & \cellcolor{gray!20}$0.18 \pm 0.08$ & $9.9 \pm 3.0$ & $1.47 \pm 0.6$ \\
& CTGAN    & $4.75 \pm 0.51$ & $3.16 \pm 0.35$ & $0.61 \pm 0.05$ & $0.559$ & \cellcolor{gray!20}$38.96 \pm 6.02$ & \cellcolor{gray!20}$0.875$ & \cellcolor{gray!20}$0.13 \pm 0.07$ & $10.8 \pm 2.2$ & $5.02 \pm 2.2$ \\
& TVAE     & $4.86 \pm 0.93$ & $1.46 \pm 0.50$ & $0.75 \pm 0.08$ & $0.826$ & \cellcolor{gray!20}$31.02 \pm 8.88$ & \cellcolor{gray!20}$0.893$ & \cellcolor{gray!20}$0.14 \pm 0.06$ & $7.1 \pm 1.3$ & $2.50 \pm 1.4$ \\
& TabPFN   & $1.61 \pm 0.12$ & $\tablebf{0.37 \pm 0.10}$ & $\tablebf{0.93 \pm 0.02}$ & $0.977$ & \cellcolor{gray!20}$\tablebf{25.20 \pm 12.01}$ & \cellcolor{gray!20}$0.961$ & \cellcolor{gray!20}$\tablebf{0.20 \pm 0.05}$ & $\tablebf{3.95 \pm 0.4}$ & $\tablebf{0.56 \pm 0.2}$ \\

\midrule
\midrule
\multirow{8}{*}{\rotatebox{90}{NN: neural network}} 
& ref.     & $0.00 \pm 0.00$ & $0.00 \pm 0.00$ & $0.85 \pm 0.07$ & $0.986$ & \cellcolor{gray!20}$27.28 \pm 12.52$ & \cellcolor{gray!20}$0.982$ & \cellcolor{gray!20}$0.15 \pm 0.09$ & $3.3 \pm 0.2$ & $0.23 \pm 0.1$ \\
\cmidrule(r){2-11}
& TabSyn   & $1.91 \pm 0.62$ & $0.57 \pm 0.23$ & $0.84 \pm 0.06$ & $0.890$ & \cellcolor{gray!20}$29.82 \pm 11.85$ & \cellcolor{gray!20}$0.875$ & \cellcolor{gray!20}$0.14 \pm 0.08$ & $4.7 \pm 0.8$ & $0.90 \pm 0.5$ \\
& STASY    & $10.25 \pm 2.80$ & $1.74 \pm 0.87$ & $0.83 \pm 0.07$ & $\tablebf{0.972}$ & \cellcolor{gray!20}$32.56 \pm 8.60$ & \cellcolor{gray!20}$0.875$ & \cellcolor{gray!20}$0.16 \pm 0.08$ & $21.3 \pm 5.8$ & $1.63 \pm 0.9$ \\
& TabDDPM  & $\tablebf{0.75 \pm 0.11}$ & $\tablebf{0.30 \pm 0.11}$ & $\tablebf{0.85 \pm 0.07}$ & $0.957$ & \cellcolor{gray!20}$29.74 \pm 11.02$ & \cellcolor{gray!20}$0.857$ & \cellcolor{gray!20}$0.16 \pm 0.09$ & $\tablebf{3.8 \pm 0.4}$ & $\tablebf{0.75 \pm 0.3}$ \\
& CoDi     & $6.72 \pm 2.73$ & $1.16 \pm 0.59$ & $0.82 \pm 0.05$ & $0.909$ & \cellcolor{gray!20}$30.50 \pm 9.89$ & \cellcolor{gray!20}$\tablebf{0.982}$ & \cellcolor{gray!20}$0.16 \pm 0.06$ & $9.5 \pm 3.9$ & $1.53 \pm 0.7$ \\
& GReaT    & $7.36 \pm 1.35$ & $1.34 \pm 0.44$ & $0.84 \pm 0.05$ & $0.943$ & \cellcolor{gray!20}$\tablebf{25.94 \pm 12.24}$ & \cellcolor{gray!20}$0.946$ & \cellcolor{gray!20}$\tablebf{0.17 \pm 0.11}$ & $9.3 \pm 2.2$ & $2.60 \pm 1.4$ \\
& CTGAN    & $4.79 \pm 0.59$ & $3.92 \pm 1.86$ & $0.58 \pm 0.04$ & $0.566$ & \cellcolor{gray!20}$39.42 \pm 5.71$ & \cellcolor{gray!20}$0.875$ & \cellcolor{gray!20}$0.13 \pm 0.06$ & $13.0 \pm 3.6$ & $7.56 \pm 3.8$ \\
& TVAE     & $5.27 \pm 1.61$ & $1.78 \pm 0.70$ & $0.69 \pm 0.08$ & $0.752$ & \cellcolor{gray!20}$31.26 \pm 9.27$ & \cellcolor{gray!20}$0.893$ & \cellcolor{gray!20}$0.15 \pm 0.06$ & $9.4 \pm 2.7$ & $4.53 \pm 2.5$ \\
& TabPFN   & $1.60 \pm 0.11$ & $0.46 \pm 0.15$ & $\tablebf{0.85 \pm 0.09}$ & $0.967$ & \cellcolor{gray!20}$26.34 \pm 11.44$ & \cellcolor{gray!20}$0.910$ & \cellcolor{gray!20}$0.15 \pm 0.10$ & $4.12 \pm 0.7$ & $0.76 \pm 0.4$ \\

    \bottomrule
    \end{tabular}}
    \vspace{-2pt}
    
\begin{flushleft}
\tiny
\textbf{Notes:} Low-order metrics~\citep{zhang2023mixed}: column/pairwise error rates of density or correlation estimation. 
Skeleton: causal skeleton level.
MEC: Markov equivalence class.
Causal direction level: SHD (structural Hamming distance), bivariate accuracy (ACC), and multivariate F1 with LiNGAM.
\setlength{\fboxsep}{1pt}  % Reduce padding
\colorbox{gray!20}{Shaded cells}: causal discovery methods are theoretically inapplicable for LG case.
Interventional task: AMAE = 100 $\times$ average mean absolute error.
\end{flushleft}
%\vspace{-10pt}
\end{table*}

\begin{table*}[ht]
    \centering
    \caption{Benchmark results under LG, LU, SG and NN causal mechanisms, avoiding variable ordering bias.}
    \vspace{-10pt}
    \label{tab:reordering_lg_lu}
    \setlength{\tabcolsep}{3pt}
    \renewcommand{\arraystretch}{0.7}
    \resizebox{0.9\linewidth}{!}{
    \begin{tabular}{c|l|cc|c|c|ccc|c|c}
    \toprule
    & \multirow{2}{*}{Model} & \multicolumn{2}{c|}{Low-order} & Skeleton & MEC & \multicolumn{3}{c|}{Causal direction level} & Intervention & Counterfact \\
    & & Col.\ ER ($\downarrow$) & Pair.\ ER ($\downarrow$) & F1 ($\uparrow$) & AUC ($\uparrow$) & SHD ($\downarrow$) & ACC ($\uparrow$) & F1 ($\uparrow$) & AMAE ($\downarrow$) & AMAE ($\downarrow$) \\
    \midrule
    \multirow{8}{*}{\rotatebox{90}{LG}} 
    & ref.     & $0.00 \pm 0.00$ & $0.00 \pm 0.00$ & $0.89 \pm 0.05$ & $0.966$ & $15.80 \pm 7.15$ & $0.571$ & $0.36 \pm 0.05 $ & $3.19 \pm 0.2$ & $0.05 \pm 0.0$ \\
    \cmidrule(r){2-11}
    & TabSyn   & $1.68 \pm 0.55$ & $1.40 \pm 2.86$ & $0.67 \pm 0.17$ & $0.840$ & \cellcolor{gray!20}23.40 $\pm$ 7.53 & \cellcolor{gray!20}0.446 & \cellcolor{gray!20}$0.31 \pm 0.08$ & $5.02 \pm 2.8$ & $1.49 \pm 2.6$ \\
    & STASY    & $10.61 \pm 5.95$ & $1.85 \pm 2.99$ & $0.64 \pm 0.17$ & $0.844$ & \cellcolor{gray!20}29.95 $\pm$ 4.87 & \cellcolor{gray!20}0.464 & \cellcolor{gray!20}0.24 $\pm$ 0.08 & $24.66 \pm 12.3$ & $2.31 \pm 3.4$ \\
    & TabDDPM  & \tablebf{0.82 $\pm$ 0.15} & $0.52 \pm 0.38$ & $0.72 \pm 0.11$ & $0.864$ & \cellcolor{gray!20}23.54 $\pm$ 7.92 & \cellcolor{gray!20}0.464 & \cellcolor{gray!20}0.28 $\pm$ 0.08 & $3.87 \pm 1.2$ & \tablebf{0.96 $\pm$ 1.4} \\
    & CoDi     & $5.26 \pm 2.41$ & $1.77 \pm 2.89$ & $0.66 \pm 0.17$ & $0.819$ & \cellcolor{gray!20}30.23 $\pm$ 5.38 & \cellcolor{gray!20}0.482 & \cellcolor{gray!20}0.20 $\pm$ 0.06 & $9.14 \pm 10.8$ & $2.57 \pm 4.4$ \\
    & GReaT    & $8.74 \pm 0.27$ & $0.99 \pm 0.39$ & $0.69 \pm 0.09$ & $0.844$ & \cellcolor{gray!20}$19.31 \pm 7.95$ & \cellcolor{gray!20}0.518 & \cellcolor{gray!20}$0.31 \pm 0.08$ & $11.50 \pm 2.3$ & $1.49 \pm 1.7$ \\
    & CTGAN    & $5.67 \pm 0.57$ & $4.17 \pm 0.91$ & $0.48 \pm 0.07$ & $0.528$ & \cellcolor{gray!20}34.71 $\pm$ 6.34 & \cellcolor{gray!20}0.643 & \cellcolor{gray!20}0.20 $\pm$ 0.08 & $16.57 \pm 5.2$ & $9.54 \pm 5.6$ \\
    & TVAE     & $4.26 \pm 1.70$ & $2.37 \pm 1.23$ & $0.58 \pm 0.06$ & $0.658$ & \cellcolor{gray!20}27.87 $\pm$ 8.31 & \cellcolor{gray!20}\tablebf{0.661} & \cellcolor{gray!20}0.23 $\pm$ 0.09 & $9.10 \pm 5.6$ & $4.48 \pm 3.0$ \\
    & TabPFN &$1.59 \pm 0.17$& \tablebf{$0.43 \pm 0.34$} &   \tablebf{$0.82\pm 0.10$} &  $\tablebf{0.897}$ & \cellcolor{gray!20}$\tablebf{18.02 \pm 9.96} $ &   \cellcolor{gray!20}$0.482 $ &  \cellcolor{gray!20}$\tablebf{0.32 \pm 0.12}$ &  $\tablebf{3.81 \pm 1.7 }$ & $ 1.01 \pm 2.4   $\\
    
    \midrule
    \midrule
    
    \multirow{8}{*}{\rotatebox{90}{LU}} 
    & ref.     & $0.00 \pm 0.00$ & $0.00 \pm 0.00$ & $0.88 \pm 0.04$ & $0.972$ & $1.54 \pm 1.58$ & $0.982$ & $0.91 \pm 0.09$ & $2.96 \pm 0.1$ & $0.02 \pm 0.0$ \\ 
    \cmidrule(r){2-11}
    & TabSyn   & $1.73 \pm 0.92$ & $0.45 \pm 0.27$ & $0.65 \pm 0.09$ & $0.749$ & $27.58 \pm 6.32$ & $0.875$ & $0.34 \pm 0.09$ & $4.19 \pm 1.4$ & $0.52 \pm 0.8$ \\
    & STASY    & $8.82 \pm 3.25$ & $1.17 \pm 0.55$ & $0.62 \pm 0.15$ & $0.857$ & $23.81 \pm 3.74$ & $\tablebf{0.964}$ & $0.40 \pm 0.11$ & $17.43 \pm 6.6$ & $0.42 \pm 0.6$ \\
    & TabDDPM  & $\tablebf{0.87 \pm 0.14}$ & $\tablebf{0.37 \pm 0.36}$ & $0.69 \pm 0.06$ & $0.759$ & $21.45 \pm 8.84$ & $0.911$ & $0.39 \pm 0.13$ & $3.96 \pm 1.4$ & $0.96 \pm 1.6$ \\
    & CoDi     & $9.19 \pm 2.52$ & $1.65 \pm 1.42$ & $0.73 \pm 0.10$ & $0.839$ & $21.57 \pm 4.93$ & $\tablebf{0.964}$ & $0.37 \pm 0.11$ & $\tablebf{3.80 \pm 0.6}$ & $0.59 \pm 0.5$ \\
    & GReaT    & $9.41 \pm 0.83$ & $0.44 \pm 0.11$ & $0.69 \pm 0.05$ & $0.808$ & $14.76 \pm 9.20$ & $0.839$ & $0.54 \pm 0.13$ & $11.09 \pm 0.8$ & $\tablebf{0.36 \pm 0.2}$ \\
    & CTGAN    & $5.75 \pm 0.97$ & $4.54 \pm 2.29$ & $0.50 \pm 0.09$ & $0.536$ & $32.23 \pm 6.08$ & $0.768$ & $0.27 \pm 0.07$ & $15.19 \pm 7.9$ & $7.92 \pm 5.4$ \\
    & TVAE     & $4.98 \pm 1.52$ & $2.35 \pm 0.75$ & $0.54 \pm 0.08$ & $0.547$ & $26.62 \pm 10.30$ & $0.714$ & $0.34 \pm 0.11$ & $10.70 \pm 3.8$ & $5.14 \pm 3.2$ \\
    & TabPFN &$1.91 \pm 0.16$& $0.48 \pm 0.14$&   \tablebf{$0.81\pm 0.09$} &  \tablebf{$0.917$} & \tablebf{$8.20 \pm 6.35$} &   $0.911 $ &  \tablebf{$0.68 \pm 0.20$} &  $4.13 \pm 2.0 $ & $ 0.39 \pm 0.6   $\\

    \midrule
    \midrule
    
    % -------------------- SG (reordering) --------------------
    \multirow{9}{*}{\rotatebox{90}{SG}} 
    & ref.     & $0.00 \pm 0.00$ & $0.00 \pm 0.00$ & $0.91 \pm 0.07$ & $0.952$ & \cellcolor{gray!20}$25.62 \pm 9.59$ & \cellcolor{gray!20}$0.964$ & \cellcolor{gray!20}$0.16 \pm 0.10$ & $3.21 \pm 0.2$ & $0.07 \pm 0.0$ \\
    \cmidrule(r){2-11}
    & TabSyn   & $1.62 \pm 0.60$ & $0.60 \pm 0.33$ & $\tablebf{0.91 \pm 0.05}$ & $0.941$ & \cellcolor{gray!20}$29.66 \pm 7.91$ & \cellcolor{gray!20}$0.929$ & \cellcolor{gray!20}$0.14 \pm 0.10$ & $4.16 \pm 1.7$ & $0.86 \pm 0.9$ \\
    & STASY    & $10.06 \pm 2.57$ & $1.71 \pm 0.83$ & $0.88 \pm 0.08$ & $0.941$ & \cellcolor{gray!20}$35.86 \pm 6.54$ & \cellcolor{gray!20}$0.946$ & \cellcolor{gray!20}$0.12 \pm 0.08$ & $23.29 \pm 6.4$ & $3.87 \pm 2.4$ \\
    & TabDDPM  & \tablebf{$0.93 \pm 0.13$} & $\tablebf{0.46 \pm 0.21}$ & $0.88 \pm 0.08$ & $0.956$ & \cellcolor{gray!20}$31.38 \pm 9.73$ & \cellcolor{gray!20}$0.893$ & \cellcolor{gray!20}$0.12 \pm 0.08$ & $\tablebf{3.62 \pm 0.5}$ & $\tablebf{0.65 \pm 0.4}$ \\
    & CoDi     & $7.46 \pm 2.43$ & $1.58 \pm 0.90$ & $0.89 \pm 0.08$ & $0.963$ & \cellcolor{gray!20}$31.12 \pm 6.73$ & \cellcolor{gray!20}$\tablebf{0.964}$ & \cellcolor{gray!20}$0.15 \pm 0.11$ & $16.20 \pm 7.5$ & $4.74 \pm 4.0$ \\
    & GReaT    & $9.53 \pm 2.58$ & $1.69 \pm 1.09$ & $0.86 \pm 0.08$ & $\tablebf{0.959}$ & \cellcolor{gray!20}$\tablebf{26.86 \pm 10.59}$ & \cellcolor{gray!20}$0.821$ & \cellcolor{gray!20}$\tablebf{0.16 \pm 0.07}$ & $12.96 \pm 3.2$ & $3.24 \pm 1.9$ \\
    & CTGAN    & $5.82 \pm 1.14$ & $3.84 \pm 0.73$ & $0.67 \pm 0.04$ & $0.830$ & \cellcolor{gray!20}$40.62 \pm 4.43$ & \cellcolor{gray!20}$0.804$ & \cellcolor{gray!20}$0.11 \pm 0.08$ & $17.97 \pm 8.7$ & $11.02 \pm 8.5$ \\
    & TVAE     & $4.65 \pm 0.67$ & $1.64 \pm 0.74$ & $0.80 \pm 0.04$ & $0.926$ & \cellcolor{gray!20}$32.34 \pm 9.16$ & \cellcolor{gray!20}$0.893$ & \cellcolor{gray!20}$0.13 \pm 0.08$ & $8.83 \pm 2.2$ & $4.06 \pm 2.2$ \\
    & TabPFN   & $1.91 \pm 0.16$ & $0.48 \pm 0.14$ & $0.90 \pm 0.08$ & $0.955$ & \cellcolor{gray!20}$28.14 \pm 9.07$ & \cellcolor{gray!20}$\tablebf{0.964}$ & \cellcolor{gray!20}$\tablebf{0.16 \pm 0.07}$ & $4.24 \pm 0.5$ & $0.78 \pm 0.4$ \\

    \midrule
    \midrule

    % -------------------- NN (reordering) --------------------
    \multirow{9}{*}{\rotatebox{90}{NN}} 
    & ref.     & $0.00 \pm 0.00$ & $0.00 \pm 0.00$ & $0.84 \pm 0.13$ & $0.962$ & \cellcolor{gray!20}$25.78 \pm 12.17$ & \cellcolor{gray!20}$0.804$ & \cellcolor{gray!20}$0.21 \pm 0.09$ & $3.17 \pm 0.1$ & $0.06 \pm 0.0$ \\
    \cmidrule(r){2-11}
    & TabSyn   & $2.46 \pm 2.26$ & $3.08 \pm 7.96$ & $0.82 \pm 0.10$ & $0.940$ & \cellcolor{gray!20}$31.26 \pm 11.41$ & \cellcolor{gray!20}$0.786$ & \cellcolor{gray!20}$0.18 \pm 0.10$ & $9.08 \pm 14.6$ & $10.74 \pm 30.7$ \\
    & STASY    & $7.69 \pm 2.34$ & $3.98 \pm 8.11$ & $0.83 \pm 0.11$ & $0.954$ & \cellcolor{gray!20}$34.96 \pm 9.57$ & \cellcolor{gray!20}$0.804$ & \cellcolor{gray!20}$0.12 \pm 0.03$ & $25.70 \pm 28.0$ & $10.12 \pm 23.2$ \\
    & TabDDPM  & \tablebf{$0.87 \pm 0.14$} & $\tablebf{0.41 \pm 0.24}$ & $\tablebf{0.84 \pm 0.12}$ & $\tablebf{0.965}$ & \cellcolor{gray!20}$31.62 \pm 10.97$ & \cellcolor{gray!20}$0.839$ & \cellcolor{gray!20}$0.16 \pm 0.07$ & $3.98 \pm 0.9$ & $0.97 \pm 1.1$ \\
    & CoDi     & $7.02 \pm 2.19$ & $3.53 \pm 7.68$ & $\tablebf{0.84 \pm 0.12}$ & $0.944$ & \cellcolor{gray!20}$32.30 \pm 12.02$ & \cellcolor{gray!20}$0.821$ & \cellcolor{gray!20}$0.16 \pm 0.07$ & $19.12 \pm 26.3$ & $8.26 \pm 16.8$ \\
    & GReaT    & $7.46 \pm 2.65$ & $3.81 \pm 7.84$ & $0.79 \pm 0.10$ & $0.924$ & \cellcolor{gray!20}$\tablebf{25.90 \pm 13.55}$ & \cellcolor{gray!20}$0.643$ & \cellcolor{gray!20}$\tablebf{0.21 \pm 0.08}$ & $16.67 \pm 24.9$ & $9.53 \pm 21.4$ \\
    & CTGAN    & $6.05 \pm 1.99$ & $6.86 \pm 6.95$ & $0.64 \pm 0.09$ & $0.788$ & \cellcolor{gray!20}$38.72 \pm 4.09$ & \cellcolor{gray!20}$0.786$ & \cellcolor{gray!20}$0.13 \pm 0.07$ & $26.12 \pm 27.2$ & $18.69 \pm 25.9$ \\
    & TVAE     & $5.53 \pm 1.59$ & $4.20 \pm 7.34$ & $0.72 \pm 0.12$ & $0.860$ & \cellcolor{gray!20}$28.74 \pm 8.26$ & \cellcolor{gray!20}$0.857$ & \cellcolor{gray!20}$0.18 \pm 0.08$ & $22.44 \pm 36.8$ & $13.16 \pm 22.3$ \\
    & TabPFN   & $1.60 \pm 0.32$ & $0.47 \pm 0.08$ & $0.77 \pm 0.14$ & $0.916$ & \cellcolor{gray!20}$30.15 \pm 10.27$ & \cellcolor{gray!20}$\tablebf{0.893}$ & \cellcolor{gray!20}$0.16 \pm 0.08$ & $\tablebf{3.92 \pm 0.6}$ & $\tablebf{0.64 \pm 0.3}$ \\
    
    \bottomrule
    \end{tabular}}
    \vspace{-6pt}
\end{table*}

\subsection{Benchmark results and discussion}\label{sec:exp_results}

We first evaluate the baseline methods on synthetic data, where ground-truth DAGs provide labels for high-order causal structure. We then assess performance on real-world data. Additional experiments (including different sample sizes, more variables, discretized variables, and extended graphs) are presented in Appendix~\ref{app:exp_details}.

\begin{figure}[t]
% \vspace{-2.5em}
  \centering
  \begin{subfigure}{0.23\textwidth}
    \includegraphics[width=\textwidth]{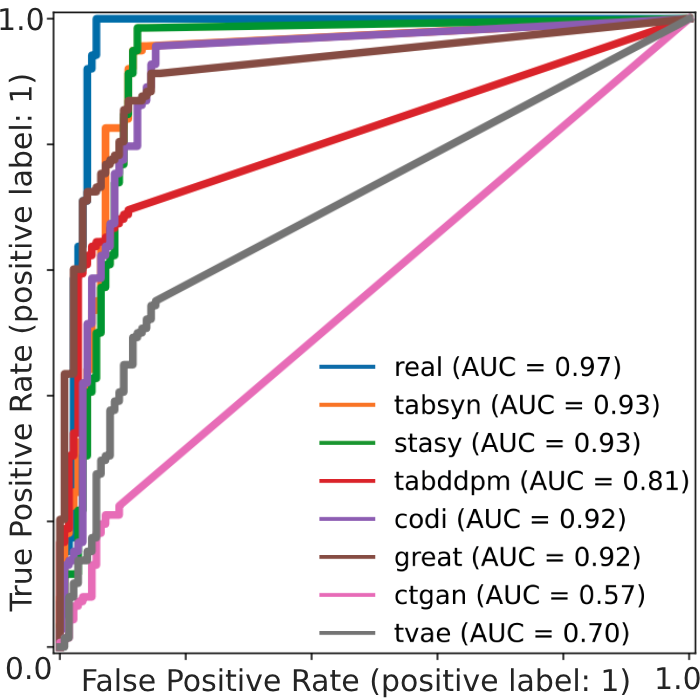}
    \caption{ROC curve (LG)}
    \label{fig:lg_roc}
  \end{subfigure}
  %\hfill
  \begin{subfigure}{0.23\textwidth}
    \includegraphics[width=\textwidth]{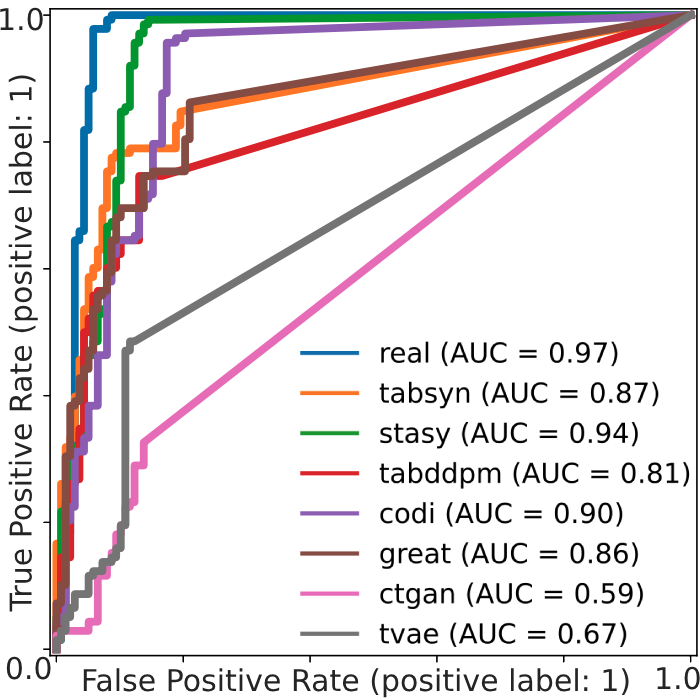}
    \caption{ROC curve (LU)}
    \label{fig:lu_roc}
  \end{subfigure}
  %\vspace{-1em}
  \caption{Benchmark on d-separations:
  ROC curves of the conditional independence test results for Markov equivalent class level evaluation.}
  \label{fig:d-sep-auc-roc}
  \vspace{-5pt}
\end{figure}

\paragraph{Results on synthetic data.}
Table~\ref{tab:combined_lg_lu_final} shows a representative subset of results under the two data-generation regimes (LG and LU). Across low-order metrics (e.g., single-column density and pairwise correlations), Transformer-based (TabSyn and TabPFN), LLM-based (GReaT) and DFM-based (TabDDPM and CoDi) models often significantly outperform older approaches like TVAE and CTGAN. However, none of the experimented models fully match reference-level high-order structures as measured by SHD, d-separation tests, or causal directionality. 
ROC curves in Figure~\ref{fig:d-sep-auc-roc} visualize the performance differences of these methods on the d-separation conditional independence tests under LG and LU settings. Notably, TabSyn achieves the second lowest pairwise correlation error (LU setting) but does not perform as well w.r.t. the causal-structure scores, confirming that low-order metrics do not reliably capture multi-variate structural fidelity.

Among the DFM-based methods, TabDDPM stands out for strong single-column density estimation and moderate high-order performance. GReaT sometimes recovers more joint structure (e.g., smaller SHD), but it can struggle on some conditional independence tasks. CoDi also consistently improves over classical generators but exhibits variability on certain causal direction metrics. Taken together, these results highlight that the best model for low-order statistics is not always the best for capturing high-order causal dependencies. TabPFN consistently performs well w.r.t. the high order metrics, which could possibly be attributed to the synthetic data it being pretrained on~\cite{tabpfn} having similarities in terms of causal structures or generation mechanisms, with the benchmarking data. 

To ensure the reported causal metrics are not driven by simple variable ordering biases in the data, we randomly permute the benchmark dataset columns before training and then restore their original order for evaluation. Table~\ref{tab:reordering_lg_lu} presents these \emph{reordered} results under LG and LU settings. The overall rankings of methods remain largely consistent, implying that state-of-the-art Transformer-based or DFM-based approaches provide better, but still imperfect, recovery of high-order structures relative to older baselines. 
Nonetheless, a clear performance gap persists between the reference data and any synthetic generator on tasks involving multivariate causal metrics.

\paragraph{Results on real-world data.}
To examine how well the benchmarked models generalize beyond synthetic settings, we evaluate them on the well-known Sachs proteomic dataset~\citep{Sachs2005-su}, a widely used benchmark in causal inference with an established ground-truth DAG. Following common practices in evaluating counterfactual inference and data imputation~\citep{almond2005costs,hill2011bayesian,geffner2022deep}, we design a missing-at-random scenario by withholding values of variable \texttt{P38}, which is known to be causally influenced by \texttt{PKC}. The baseline models are trained on the observed training partition and then used to generate synthetic datasets for imputing \texttt{P38}.

\begin{table*}[t!]
    \centering     
    \caption{\small Imputation results for the $P38$ feature in the Sachs dataset~\citep{Sachs2005-su}, evaluated using low-order metrics from~\citet{zhang2023mixed}, mean absolute error (MAE) and mean squared error (MSE), across XGBoost~\citep{DBLP:journals/corr/ChenG16}, MLP, and TabPFN~\citep{tabpfn}.}
    \vspace{-10pt}
    \label{tab:sachs-main}
    \setlength{\tabcolsep}{3pt}
     \resizebox{0.95\textwidth}{!}{
    \begin{tabular}{l|cc|cc|cc|cc}
    \bottomrule
& \multicolumn{2}{c|}{Low-order} & \multicolumn{2}{c|}{XGBoost} & \multicolumn{2}{c|}{MLP} & \multicolumn{2}{c}{TabPFN (i)}  \\ 
& Col.\ ER ($\downarrow$) & Pair.\ ER ($\downarrow$) & MAE ($\downarrow$) & MSE($\downarrow$) & MAE ($\downarrow$) & MSE($\downarrow$) & MAE ($\downarrow$) & MSE($\downarrow$) \\ \midrule
TabSyn	&$\hphantom{0}2.54\!\pm\!0.51$& $2.30\!\pm\!0.71$ & $181.46\!\pm\!43.16$ & $2.38\mathrm{e}{+}05\!\pm\!1.24\mathrm{e}{+}05$ & $31.84\!\pm\!4.93$&$2.00\mathrm{e}{+}04\!\pm\!2.09\mathrm{e}{+}03$ & $223.84\!\pm\!33.52$&$4.39\mathrm{e}{+}05\!\pm\!1.51\mathrm{e}{+}05$\\
    STASY	&$\hphantom{0}12.54\!\pm\!4.72$&$3.08\!\pm\!1.43$ & $\tablebf{106.84\!\pm\!18.03}$ & $7.89\mathrm{e}{+}04\!\pm\!3.96\mathrm{e}{+}04$ & $28.12\!\pm\!2.96$&$1.78\mathrm{e}{+}04\!\pm\!1.02\mathrm{e}{+}03$ & $160.28\!\pm\!34.25$&$2.18\mathrm{e}{+}05\!\pm\!1.15\mathrm{e}{+}05$\\
    TabDDPM &$\hphantom{0}\tablebf{2.66\!\pm\!0.28}$&$\tablebf{1.42\!\pm\!0.44}$ & $108.56\!\pm\!11.35$ & $\tablebf{7.51\mathrm{e}{+}04\!\pm\!1.82\mathrm{e}{+}04}$ & $\tablebf{28.04\!\pm\!1.32}$&$\tablebf{1.70\mathrm{e}{+}04\!\pm\!1.01\mathrm{e}{+}03}$ & $151.84\!\pm\!14.24$&$1.88\mathrm{e}{+}05\!\pm\!4.08\mathrm{e}{+}04$\\
    CoDi	&$\hphantom{0}37.47\!\pm\!3.57$&$10.03\!\pm\!2.04$ & $315.56\!\pm\!184.39$ & $4.96\mathrm{e}{+}05\!\pm\!4.26\mathrm{e}{+}05$  & $46.89\!\pm\!16.63$&$2.36\mathrm{e}{+}04\!\pm\!1.18\mathrm{e}{+}04$ & $\tablebf{131.71\!\pm\!82.44}$&$\tablebf{9.34\mathrm{e}{+}04\!\pm\!8.30\mathrm{e}{+}04}$\\
    GReaT   &$\hphantom{0}16.07\!\pm\!0.23$&$3.27\!\pm\!0.49$ & $275.58\!\pm\!43.68$ & $4.67\mathrm{e}{+}05\!\pm\!1.35\mathrm{e}{+}05$ & $43.54\!\pm\!7.21$&$3.75\mathrm{e}{+}04\!\pm\!5.86\mathrm{e}{+}03$ & $217.61\!\pm\!22.92$&$3.39\mathrm{e}{+}05\!\pm\!8.66\mathrm{e}{+}04$\\
    CTGAN	&$\hphantom{0}14.57\!\pm\!1.45$&$8.54\!\pm\!0.64$ & $335.08\!\pm\!32.11$ & $7.35\mathrm{e}{+}05\!\pm\!8.74\mathrm{e}{+}04$ & $45.83\!\pm\!7.13$&$4.41\mathrm{e}{+}04\!\pm\!7.97\mathrm{e}{+}03$ & $363.49\!\pm\!29.36$&$8.71\mathrm{e}{+}05\!\pm\!8.33\mathrm{e}{+}04$\\
    TVAE	&$\hphantom{0}13.39\!\pm\!3.00$&$5.69\!\pm\!1.31$ & $283.05\!\pm\!43.09$ & $6.42\mathrm{e}{+}05\!\pm\!1.46\mathrm{e}{+}05$ & $28.11\!\pm\!1.80$&$2.93\mathrm{e}{+}04\!\pm\!9.70\mathrm{e}{+}03$ & $228.73\!\pm\!39.58$&$4.55\mathrm{e}{+}05\!\pm\!1.03\mathrm{e}{+}05$\\
    TabPFN (g)	&$\hphantom{0}4.52\!\pm\!0.21$&$8.17\!\pm\!0.73$ & $261.23\!\pm\!79.17$ & $8.88\mathrm{e}{+}05\!\pm\!7.00\mathrm{e}{+}05$ & $72.51\!\pm\!9.86$&$6.23\mathrm{e}{+}04\!\pm\!2.19\mathrm{e}{+}04$ & $181.81\!\pm\!33.51$&$3.94\mathrm{e}{+}05\!\pm\!1.79\mathrm{e}{+}05$\\
\toprule
    \end{tabular} }
    \vspace{-10pt}
\end{table*}

We evaluate the imputation quality using three regression models: XGBoost~\citep{DBLP:journals/corr/ChenG16}, a shallow MLP, and TabPFN~\citep{tabpfn} both for generation (g) and for imputation (i); they are trained solely on the synthetic data and tested on held-out real observations. 
As shown in Table~\ref{tab:sachs-main}, performance varies considerably. TabDDPM consistently ranks among the top performers, achieving the lowest MSE on both XGBoost and MLP regressors. 
STASY yields the best MAE with XGBoost, and CoDi achieves the strongest TabPFN (i) results. TabPFN (g) performs poorly in comparison to its strong performance in the fully synthetic setting in Table ~\ref{tab:combined_lg_lu_final}, the performance discrepancy could possibly be attributed to the Sachs dataset or it's causal structure being more dissimilar from the datasets and structures TabPFN was pretrained on. Furthermore the issues raised in \citep{reisach2021beware} highlight how artifacts in synthetic causal data can induce inductive biases that do not transfer to real-world settings, potentially leading to an effective prior mismatch when TabPFN is applied to the Sachs dataset. Regardless of its reason, this discrepancy in performance further indicates that high-order metrics alone are not sufficient to address generalisability to real-world settings, but rather the high- and low-order metrics should be used in conjunction when evaluating tabular synthesis models, especially when real-world data is concerned.
Notably, these results correlate with models' downstream counterfactual inference scores under synthetic benchmarks (Appendix~\ref{app:exp_details}, Table~\ref{tab:eva_causal_inference}), indicating that high-order causal fidelity does translate into real-world utility for decision-support tasks.

Despite these observations, no model dominates across all metrics and learners, suggesting that current methods remain limited in consistently modeling the full structural complexity of tabular data. This highlights the importance of high-order benchmarking as a tool to guide improvements in causal-aware synthetic data generation.

\section{Conclusion}

We present a benchmark framework designed to systematically evaluate tabular synthesis models with respect to their capability of capturing high-order structural causal information. By introducing causal graphs as explicit and meaningful prior knowledge, we categorize causal dependencies into three hierarchical levels, facilitating precise benchmark tasks and metrics. To overcome challenges associated with ground-truth availability, we develop synthetic benchmark datasets complemented by causal discovery techniques, enabling robust evaluation across multiple causal dimensions. Our framework effectively distinguishes the strengths and limitations of contemporary synthesis models, providing insights for model improvement. Notably, our evaluations reveal that current state-of-the-art methods, although excelling in capturing lower-order statistical properties, still exhibit significant gaps in modeling high-order causal structures. These results highlight the necessity for developing synthesis methods that inherently incorporate structural priors or enforce causal constraints during training.  We emphasize that CauTabBench is a controlled diagnostic benchmark for causal fidelity rather than a one-number universal model selector; it complements low-order and downstream evaluations. Ultimately, our benchmark advances methodological rigor in tabular data synthesis and highlights its relevance to high-stakes domains where causal fidelity is essential. A natural extension is to define reusable causal-graph templates i.e. confounder, mediator, collider, and domain-specific motifs. In order to scale the benchmark to higher-dimensional and practitioner-facing settings. Future work and limitations are further discussed in Appendix~\ref{app:limitation}.

\section*{Ethics Statement}
The proposed benchmark framework focuses exclusively on synthetic and publicly available datasets and does not involve human subjects, personal data, or sensitive information. All benchmark datasets used in this study were either generated via predefined causal graphs under controlled simulation processes or drawn from established open datasets that are licensed for research purposes (e.g., UCI Machine Learning Repository, DELVE repository). No private or proprietary data were used.

Potential negative societal impacts include the possibility of misinterpretation of benchmark results if applied without consideration of the assumptions underlying causal discovery methods. As discussed in our limitations (Appendix~\ref{app:limitation}), violations of these assumptions may affect transferability to real-world scenarios, and fairness-related concerns are beyond the current scope of this framework. We emphasize that our benchmark is intended as a research tool to advance the methodological rigor of tabular data synthesis, rather than as a direct application pipeline. We provide full access to all experimental code, data, and documentation to foster transparency, accountability, and responsible research practices.

\section*{Reproducibility Statement}
We have taken extensive measures to ensure the reproducibility of our results. The benchmark framework, including code, data generation procedures, and documentation, is publicly available at \url{https://github.com/TURuibo/CauTabBench}, and the corresponding datasets can be accessed at \url{https://doi.org/10.7910/DVN/EB0KCO}. Detailed descriptions of benchmark dataset generation, causal mechanisms, and noise models are presented in Section~\ref{sec:benchmark_data}. Evaluation metrics for high-order structural causal information, conditional independence, and causal direction are formally defined in Section~\ref{sec:benchmark_task}, with additional methodological details provided in Appendix~\ref{app:causal_discovery}.

Our experimental settings, including baseline implementations, training configurations, and evaluation procedures, are described in Section~\ref{sec:exp_setting} and Appendix~\ref{app:exp_details}. Results are reported with mean and standard deviation across multiple random seeds and bootstrapped datasets, reducing variance due to randomness. To mitigate variable-ordering bias, we conducted experiments with randomized column orders (Appendix~\ref{app:exp_details_var_order_bias}). We also include experiments on real-world datasets to assess external validity (Appendix~\ref{app:exp_realworld_datasets}). Together, these measures ensure that independent researchers can reproduce and verify our findings with the provided resources.

\begin{acks}
The benchmarking results of large language foundation models were enabled by resources provided by the National Academic Infrastructure for Supercomputing in Sweden (NAISS), partially funded by the Swedish Research Council through grant agreement no.~2022-06725.
\end{acks}
\newpage
\bibliography{ref}
\bibliographystyle{ACM-Reference-Format}

\newpage

% Start of the table of contents
\setcounter{tocdepth}{0} % Set tocdepth to 0 to exclude sections before this point
\addtocontents{toc}{\protect\setcounter{tocdepth}{3}} % Set tocdepth to 3 (or any other value you prefer) to include sections, subsections, and subsubsections after this point

\appendix
\section{Brief Introduction of Causal Discovery} \label{app:causal_discovery}
Causal discovery aims to determine causal relationships purely based on observational data by leveraging their statistical properties under proper assumptions~\citep{spirtes2000causation,peters2017elements}. 
The methodology of causal discovery can be characterized into constraint-based, score-based, and Functional Causal Model (FCM)-based methods~\citep{glymour2019review}. \textit{Constraint-based methods}, such as PC and FCI algorithms, apply conditional independence tests to each pair of variables and infer causal skeletons and causal directions based on certain rules~\citep{spirtes2000causation}. \textit{Score-based methods }~\cite{chickering2002optimal,huang2018generalized} formulate causal discovery as an optimization problem and optimize score functions by searching in the space of DAGs. Many deep learning-based methods~\citep{zheng2018dags,ng2022convergence} can be considered as score-based ones. Under Markov and causal faithfulness assumptions, constraint-based and score-based methods identify causal graphs up to certain equivalence classes. 
For example, suppose that there are no unknown confounders, PC algorithm identifies the Markov equivalence classes of causal DAGs as CPDAGs, and the results of score-based method GES~\citep{chickering2002optimal} also converge to Markov equivalence classes. 
To further identify causal relationships in the same equivalence class, \textit{FCM-based methods} impose additional assumptions of functional classes and distributions on the data generation processes. The most flexible and identifiable FCM is proposed by~\cite{hoyer2008nonlinear}. Common FCM-based methods~\citep{shimizu2006linear,hoyer2008nonlinear} are based on the bivariate case determining the cause and the effect between two variables. For example, one can fit two FCMs in different directions and select the one with large likelihood on the observational data. There are also multivariate FCM-based methods, such as LiNGAM~\citep{shimizu2006linear} that assumes that the FCM is a linear non-Gaussian model.

\paragraph{Assumptions of causal discovery methods.}
Throughout the paper, we frequently mention proper assumptions without explicitly stating them. This is because assumptions for different causal discovery methods are different and some of them involve specific and technical details such that consistently stating them would make the paper unnecessarily complicated and more difficult to follow. For the sake of brevity and clarity, we use "under proper assumptions" in general. The involved assumptions on the causal skeleton level: 
\begin{itemize}
    \item The identifiability conditions of PC algorithm~\citep{spirtes2000causation,peters2017elements}.
\end{itemize}

The involved assumptions on the causal direction level:
\begin{itemize}
    \item The identifiability conditions of additive noise models~\citep{hoyer2008nonlinear};
    \item The identifiability conditions of post-nonlinear models~\citep{zhang2009identifiability};
    \item The identifiability conditions of LiNGAM~\citep{shimizu2006linear}.
\end{itemize}

\paragraph{Potential negative societal impacts.}
Since real-world scenarios are always violating the proper assumptions in different ways, the benchmarking results need to be carefully interpreted together with the causal discovery assumptions. Considering the potential violation of the assumptions for specific applications is necessary for a proper usage of the benchmark framework. Although we conduct experiments on real-world data, as shown in Tables~\ref{tab:sachs-main} and~\ref{tab:real} that show an indication of the transferability of the framework to real-world settings to some extent, we advice against concluding that this transferability always holds, as our experiments on this are not exhaustive and merely a first attempt at bridging this gap. While the proposed framework accounts for certain biases (e.g.,~variable ordering bias) in evaluating baseline models, broader concerns such as fairness are beyond the current scope and represent valuable directions for future work. We suggest using separate evaluation and mitigation techniques to complement the proposed framework in order to address this, prior to any application of the baseline models. 

\paragraph{Best Practices in using causal discovery methods for benchmarking on high-order causal structural information.}
Benchmark datasets with linear relationships and simple distributions (e.g., Gaussian and uniform distributions) are good enough for distinguishing models in terms of capturing causal information. Moreover, the conditional independence tests in constraint-based causal discovery methods can be efficiently applied to datasets with linear relationships; whereas, the kernel-based tests for the datasets with nonlinear relationships are only feasible when the sample size is around $1000$.
As for the application of bivariate causal discovery methods, we find that the methods without requiring training models are more efficient for the purpose of evaluation and can provide a reasonable evaluation results. 
And in general, their results do not vary a lot, and do not require bootstrapping for an error bar.

\section{Limitations and Future Works} \label{app:limitation}
This work aims at establishing the foundation of benchmarking tabular synthesis models on causal information and future works can consider to further address some limitations. 
CauTabBench is designed as a controlled diagnostic benchmark for causal fidelity. Its central claim is that, on data with known high-order causal structure, the framework reveals whether a synthesis model preserves that structure. It is not intended as a universal one-number model selector across all real-world tabular settings, and the limitations below should be read in that light.
We conducted initial experimentation with respect to discrete data in Table~\ref{tab:discrete}, but we suggest to expand the benchmark framework with more generation and discretisation processes in addition to mixed data types while satisfying Conditions (i) and (ii) while including more baseline methods. 
Current benchmark datasets are based mainly on causal mechanisms $LG$, $LU$, $SG$, and $NG$, and our experimental setup uses $N=10$ or $20$ variables which could be easily expanded as $N$ is a configurable parameter in our benchmark framework. Although for the purpose of this paper, it is sufficient with our current setting, it may not be suitable for specific applications or other use cases. 
Additionally, since this work is based on a general consideration of causal relationships without a specific downstream application, we only choose general benchmark tasks and metrics based on causal discovery algorithms. It is worthwhile to design more task-specific benchmark tasks and metrics when there are specific downstream applications of synthetic tabular data.
Moreover, the current experiment setting is too ideal compared with real-world scenarios. More factors can be considered so that benchmark datasets are close to real-world cases. For example, we currently assume that all the variables of a causal graph are known; however, it is not always the case in real-world scenario. And future works can consider to include unknown confounders.
Last but not least, our evaluation is limited by the assumptions of causal discovery algorithms, because causal information used for evaluation is not the direct outputs of tabular synthesis models but is extracted by causal discovery methods. 
For example, we use causal DAGs instead of causal graphs because common causal discovery methods rely on the DAG assumption and the purpose of the work is to evaluate synthesis models, for which the validity is more important than the flexibility of structural causal models. And the mixed data type of continuous and discrete data is not included in the benchmark because there is still lack of studies in causal discovery on such data. This also motivates the research studies in causal discovery domain, which have potentials for modelling tabular data and evaluating synthesis models.

Furthermore, as demonstrated in~\citep{reisach2021beware}, existing graph simulation methodologies, including those applied in this study, exhibit notable limitations. In particular, the widely used Erd\H{o}s-R\'enyi model for generating random DAGs does not produce a uniform distribution over the space of possible graph structures. This highlights the importance of further examining the transferability of the framework to real-world scenarios and supports the investigation of alternative graph simulation strategies that address the limitations identified in~\citep{reisach2021beware}.

\section{More Experiments}
\label{app:exp_details}
\subsection{Hardware, datasets, software, and implementation}
\paragraph{Hardware.}
We used one NVIDIA RTX 2080 Ti for the benchmarking results for the non-reordered benchmarking using 10 variables. A Google Cloud g2-standard-32 virtual machine with an NVIDIA-L4 accelerator was used for the discretized, additional variables and Sachs benchmarking.

\paragraph{Implementation of the benchmark framework.} Our benchmark framework is available at \url{https://github.com/TURuibo/CauTabBench}.

\paragraph{Baseline methods.}
Baseline methods are implemented based on the repository 
\url{https://github.com/amazon-science/tabsyn} of~\citep{zhang2023mixed}. 
As shown in Table~\ref{tab:low}, we evaluate the synthetic datasets with the metrics in~\citep{zhang2023mixed} as the reproduced results for a sanity check and a reference for other works. For the TabPFN \citep{tabpfn} generation and imputation settings we use the official implementations \url{https://github.com/PriorLabs/TabPFN}, \url{https://github.com/priorlabs/tabpfn-extensions}.
We used the training configurations provided for the "Magic" dataset~\citep{misc_magic_gamma_telescope_159} to train baseline methods.

\paragraph{Benchmark dataset generation.}
\leavevmode\\
We modify~\href{https://fentechsolutions.github.io/CausalDiscoveryToolbox/html/index.html}{\texttt{CausalDiscoveryToolbox}} for generating benchmark datasets with randomly generated causal DAGs. For demonstration, we used the configuration, variable types, number of variables, and sample size of a real-world dataset~\citep{misc_magic_gamma_telescope_159}, which is also used for our evaluation of real-world datasets. 10 causal DAGs are randomly generated and each has 10 nodes representing continuous variables and 1 node representing a binary variable. The binary variable in~\citep{misc_magic_gamma_telescope_159} is the classification target variable. For each causal DAG, we generate benchmark datasets of which the sample size is 17,117. We find that there is a lack of implementation for causal discovery methods in the presence of mixed data types; hence, we generate the binary variable independent of all other variables. In this way, we train the baseline methods on 11 variables and evaluate on 10 continuous variables by dropping the binary one, reducing the binary variable's influence on the evaluation. We used random seeds from 100 to 109 to generate the 10 causal graphs and their corresponding benchmark datasets.

\paragraph{Modifications to \texttt{CausalDiscoveryToolbox}.}
We extended \texttt{CausalDiscoveryToolbox} in two benchmark-specific ways: (1) a random acyclic graph generator tailored to our benchmark settings, with configurable graph size and bounded in-degree, and (2) causal mechanisms implementing the LG/LU/SG/NG generation processes described in Section~\ref{sec:benchmark_data}, together with the discretization step used in the discrete-data experiments. These changes are implemented in \texttt{utils/acyclic\_graph\_generator.py} and \texttt{utils/causal\_mechanisms.py} of the released code.

\paragraph{Causal discovery and inference methods for evaluation.}
\leavevmode\\
\href{https://github.com/py-why/causal-learn}{\texttt{causal-learn}}~\citep{zheng2024causal}
is used to evaluate the causal skeleton level and Markov equivalent class level. The conditional independence test is Fisher's-Z tests for datasets with linear relationships and is kernel conditional independence tests for datasets with nonlinear relationships. \href{https://github.com/FenTechSolutions/CausalDiscoveryToolbox}{\texttt{CausalDiscoveryToolbox}} is used for the causal direction level. And \href{https://github.com/py-why/dowhy}{\texttt{DoWhy}}~\citep{dowhy,dowhy_gcm} is used for the evaluation on the causal inference downstream tasks with additive noise models.

\subsection{Scaling to 20-node graphs}\label{app:20node}
To investigate how DAG size affects the framework, we extend the experiments to 20 nodes (Table~\ref{tab:20}). Compared with the 10-node setting, the 20-node experiment does not overturn our conclusions. The benchmark still clearly separates reference data from synthetic generators, and the central qualitative finding remains unchanged: strong low-order fidelity does not reliably translate into strong recovery of high-order causal structure.

\subsection{Discrete-variable experiments}
To investigate whether the framework generalizes to discrete data, we apply the proposed discretization procedure (Section~\ref{sec:benchmark_data}) to the generated datasets prior to training the baseline models and evaluating them using the framework (Table~\ref{tab:discrete}).

\paragraph{Real-world datasets used for the evaluation.}
We used 4 real-world datasets for the evaluation, which are suitable for our evaluation on the high-order structural causal information. Because they are based on linear relationships and continuous variables with a causal semantic context. They are Beijing~\citep{misc_beijing}, Magic~\citep{misc_magic_gamma_telescope_159}, \href{https://www.openml.org/search?type=data&id=821&sort=runs&status=active}{House}~\citep{vanschoren2014openml}, and Parkinsons~\citep{misc_parkinsons_telemonitoring_189}.
These datasets are licensed under a Creative Commons Attribution 4.0 International (CC BY 4.0) license.

\begin{table*}[!t]
    \centering     
    \caption{Benchmark on low-order statistics. 
    Values are mean and standard deviation of metric values (error rate (\%) of single column density, error rate (\%) of pair-wise correlation score, $\alpha-$precision, $\beta-$recall) over $10$ random causal DAGs.}\label{tab:low}
    \begin{subtable}[b]{0.48\linewidth}
        \centering        
    \caption{Linear Gaussian}
    % \label{tab:lg}
    \resizebox{0.99\columnwidth}{!}{
    \begin{tabular}{l|llll}
    \hline
    & Col.  	 & Pair. 	 & $\alpha$-precision 	& $\beta$-recall \\ \hline 
TabSyn	 &$2.11 \pm 1.08$& $0.61 \pm 0.21$ &$98.41 \pm 1.39$&$49.34 \pm 9.09$\\ 
STASY	 &$12.42 \pm 3.27$& $1.31 \pm 0.81$&$93.93 \pm 3.92$&$43.67 \pm 5.49$\\ 
TabDDPM	 &$0.69 \pm 0.12$& $0.62 \pm 0.54$ &$99.34 \pm 0.14$&$50.00 \pm 0.40$\\ 
CoDi	 &$4.42 \pm 0.83$& $0.80 \pm 0.43$ &$84.60 \pm 3.13$&$58.50 \pm 2.02$\\ 
GReaT	 &$8.41 \pm 0.73$& $0.76 \pm 0.26$ &$81.55 \pm 6.78$&$51.49 \pm 1.18$\\ 
CTGAN	 &$4.70 \pm 0.78$& $3.76 \pm 0.63$ &$88.29 \pm 4.09$&$13.23 \pm 9.09$\\ 
TVAE	 &$4.51 \pm 1.64$& $1.93 \pm 0.64$ &$87.17 \pm 9.68$&$30.77 \pm 10.16$\\ 
TabPFN	 &$1.63 \pm 0.17$& $0.30 \pm 0.09$&$93.94 \pm 1.55$&$51.97 \pm 0.24$\\ 
\hline
    \end{tabular}}
    \end{subtable}    
    \begin{subtable}[b]{0.48 \linewidth}
    \centering        
    \caption{Linear uniform}
    % \label{tab:lu}
    \resizebox{0.99\columnwidth}{!}{
    \begin{tabular}{l|llll}
    \hline
    & Col.  	 & Pair. 	 & $\alpha$-precision 	& $\beta$-recall \\ \hline 
TabSyn	 &$1.88 \pm 0.87$& $0.45 \pm 0.28$&$98.40 \pm 1.01$&$49.81 \pm 0.95$\\ 
STASY	 &$11.90 \pm 5.72$& $1.18 \pm 0.59$&$93.71 \pm 5.70$&$47.37 \pm 2.62$\\ 
TabDDPM	 &$0.98 \pm 0.63$& $1.07 \pm 2.27$&$99.24 \pm 0.35$&$41.06 \pm 18.73$\\ 
CoDi	 &$8.01 \pm 1.55$& $1.75 \pm 1.52$&$66.48 \pm 2.37$&$59.84 \pm 2.56$\\ 
GReaT	 &$9.56 \pm 0.73$& $0.56 \pm 0.22$&$96.22 \pm 0.84$&$46.17 \pm 1.09$\\ 
CTGAN	 &$4.71 \pm 0.50$& $3.52 \pm 0.50$&$92.35 \pm 2.37$&$8.06 \pm 8.69$\\ 
TVAE	 &$6.33 \pm 1.69$& $2.52 \pm 1.17$&$92.78 \pm 3.83$&$14.99 \pm 10.87$\\ 
TabPFN	 &$2.04 \pm 0.40$& $0.30 \pm 0.10$&$98.25 \pm 0.74$&$50.10 \pm 0.43$\\

\hline
    \end{tabular}}
    \end{subtable}  
    \begin{subtable}[b]{0.48 \linewidth}
    \centering        
    \caption{Sigmoid Gaussian}
    % \label{tab:sg}
    \resizebox{0.99\columnwidth}{!}{
    \begin{tabular}{l|llll}
    \hline
    & Col.  	 & Pair. 	 & $\alpha$-precision 	& $\beta$-recall \\ \hline 
TabSyn	 &$1.83 \pm 0.84$& $0.49 \pm 0.20$&$98.64 \pm 1.05$&$49.37 \pm 0.64$\\ 
STASY	 &$12.48 \pm 5.13$& $1.52 \pm 0.43$&$92.77 \pm 6.81$&$44.83 \pm 3.78$\\ 
TabDDPM	 &$0.85 \pm 0.14$& $0.40 \pm 0.30$&$99.18 \pm 0.40$&$49.83 \pm 0.69$\\ 
CoDi	 &$5.96 \pm 2.53$& $1.14 \pm 0.29$&$92.86 \pm 4.46$&$61.66 \pm 2.28$\\ 
GReaT	 &$8.51 \pm 2.03$& $1.45 \pm 0.72$&$86.12 \pm 5.82$&$50.65 \pm 2.58$\\ 
CTGAN	 &$4.75 \pm 0.51$& $3.16 \pm 0.35$&$89.44 \pm 4.16$&$23.01 \pm 6.96$\\ 
TVAE	 &$4.86 \pm 0.93$& $1.46 \pm 0.50$&$89.70 \pm 5.98$&$41.38 \pm 6.49$\\ 
TabPFN	 &$1.61 \pm 0.12$& $0.37 \pm 0.10$&$94.16 \pm 1.05$&$52.14 \pm 0.64$\\

\hline
    \end{tabular}}
    \end{subtable}   
    \begin{subtable}[b]{0.48 \linewidth}
    \centering        
    \caption{Neural network Gaussian}
    % \label{tab:nn}
    \resizebox{0.99\columnwidth}{!}{
    \begin{tabular}{l|llll}
    \hline
    & Col.  	 & Pair. 	 & $\alpha$-precision 	& $\beta$-recall \\ \hline 
TabSyn	 &$1.91 \pm 0.62$& $0.57 \pm 0.23$&$98.41 \pm 1.19$&$48.93 \pm 0.75$\\ 
STASY	 &$10.25 \pm 2.80$& $1.74 \pm 0.87$&$92.57 \pm 6.11$&$47.55 \pm 2.99$\\ 
TabDDPM	 &$0.75 \pm 0.11$& $0.30 \pm 0.11$&$99.16 \pm 0.32$&$50.27 \pm 0.87$\\ 
CoDi	 &$6.72 \pm 2.73$& $1.16 \pm 0.59$&$86.91 \pm 6.90$&$58.71 \pm 3.76$\\ 
GReaT	 &$7.36 \pm 1.35$& $1.34 \pm 0.44$&$87.40 \pm 5.86$&$48.15 \pm 4.07$\\ 
CTGAN	 &$4.79 \pm 0.59$& $3.92 \pm 1.86$&$89.49 \pm 3.95$&$11.13 \pm 10.92$\\ 
TVAE	 &$5.27 \pm 1.61$& $1.78 \pm 0.70$&$90.38 \pm 6.74$&$22.04 \pm 15.71$\\ 
TabPFN	 &$1.60 \pm 0.11$& $0.46 \pm 0.15$&$94.97 \pm 1.40$&$51.81 \pm 0.57$\\ 

\hline
    \end{tabular}}
    \end{subtable}   

\end{table*}

\begin{table*}[!t]
    \centering     
    \caption{Discretized data (see Section~\ref{sec:benchmark_data} in main paper) Results. \ref{tab:discrete-lo} 
    Values are mean and standard deviation of metric values (error rate (\%) of single column density, error rate (\%) of pair-wise correlation score, $\alpha-$precision, $\beta-$recall) over $10$ random causal DAGs. \ref{tab:discrete-skel} Benchmark on causal skeletons. Values are mean and standard deviation of metric values
(SHD, F1 score, recall, and precision) over 10 random causal DAGs. Each metric value on a causal
graph is the average value over 5 bootstrapping datasets. \ref{tab:discrete-high} Causal direction level and Interventional downstream tasks.}\label{tab:discrete}
    \begin{subtable}[b]{0.54\linewidth}
        \centering        
    \caption{Low-Order}
    \label{tab:discrete-lo}
    \resizebox{0.95\columnwidth}{!}{
    \begin{tabular}{l|llll}
    \hline
        & Col.  	 & Pair. 	 & $\alpha$-precision 	& $\beta$-recall \\ \hline 
TabSyn	 &$2.14 \pm 0.66$& $4.94 \pm 0.78$&$99.00 \pm 0.57$&$70.60 \pm 0.65$\\ 
STASY	 &$24.29 \pm 5.34$& $34.83 \pm 7.33$&$88.49 \pm 5.35$&$55.99 \pm 4.80$\\ 
TabDDPM	 &$1.05 \pm 0.09$& $4.62 \pm 0.05$&$99.33 \pm 0.27$&$68.82 \pm 0.45$\\ 
CoDi	 &$23.14 \pm 1.47$& $37.60 \pm 1.57$&$54.33 \pm 8.70$&$47.66 \pm 1.35$\\ 
GReaT	 &$1.17 \pm 0.10$& $4.64 \pm 0.08$&$98.47 \pm 0.46$&$69.22 \pm 0.44$\\ 
CTGAN	 &$22.06 \pm 0.83$& $35.04 \pm 0.89$&$92.91 \pm 3.97$&$50.85 \pm 1.35$\\ 
TVAE	 &$83.96 \pm 16.89$& $95.38 \pm 10.63$&$18.02 \pm 7.09$&$1.61 \pm 4.68$\\ 
TabPFN	 &$18.11 \pm 0.61$& $25.76 \pm 0.90$&$90.22 \pm 5.22$&$60.30 \pm 1.05$\\ 

\hline
    \end{tabular}}
    \end{subtable}   
    \begin{subtable}[b]{0.45\linewidth}
        \centering        
    \caption{Skel}
    \label{tab:discrete-skel}
    \resizebox{0.95\columnwidth}{!}{
    \begin{tabular}{l|llll}
    \hline
	& Adj  	 & F1 	 & Precision 	& Recall \\ \hline
ref.& $26.30\pm 5.99$	 & $0.23\pm 0.05$	 & $0.15\pm 0.06$	 & $0.21\pm 0.09$\\ 
\hline
TabSyn	 & $27.46\pm 4.82$	 & $0.22\pm 0.06$	 & $0.17\pm 0.06$	 & $0.20\pm 0.10$\\ 
STASY	 & $35.30\pm 4.08$	 & $0.23\pm 0.04$	 & $0.30\pm 0.05$	 & $0.19\pm 0.06$\\ 
TabDDPM	 & $26.46\pm 3.41$	 & $0.21\pm 0.03$	 & $0.14\pm 0.06$	 & $0.19\pm 0.11$\\ 
CoDi	 & $30.96\pm 5.15$	 & $0.26\pm 0.10$	 & $0.30\pm 0.11$	 & $0.23\pm 0.12$\\ 
GReaT	 & $25.96\pm 4.76$	 & $0.21\pm 0.02$	 & $0.16\pm 0.05$	 & $0.21\pm 0.08$\\ 
CTGAN	 & $53.82\pm 6.68$	 & $0.30\pm 0.06$	 & $0.67\pm 0.16$	 & $0.20\pm 0.05$\\ 
TVAE	 & $69.04\pm 9.87$	 & $0.33\pm 0.09$	 & $0.96\pm 0.13$	 & $0.20\pm 0.06$\\ 
TabPFN	 & $28.94\pm 4.29$	 & $0.22\pm 0.07$	 & $0.21\pm 0.08$	 & $0.20\pm 0.10$\\ 

\hline
    \end{tabular}}
    \end{subtable}    
    \\
\vspace{0.5em}
    \begin{subtable}[b]{0.48\linewidth}
        \centering        
    \caption{High-Order}
    \label{tab:discrete-high}
    \resizebox{0.95\columnwidth}{!}{
    \begin{tabular}{l|lllll}
    \hline
	& MEC AUC  	 & RECI 	 & IGCI & CDS & Intv   \\ \hline
ref.& $0.500$	 & $0.518$	 & $0.482$& $0.464$ & $9.50 \pm 0.3$\\ 
\hline
TabSyn	 & $0.500$	 & $0.607$	 & $0.393$	& $0.589$	 & $10.37 \pm 0.6$	\\ 
STASY	 & $0.500$	 & $0.643$	 & $0.357$	& $0.679$	 & $55.01 \pm 15.3$	\\ 
TabDDPM	 & $0.501$	 & $0.500$	 & $0.500$	& $0.429$	 & $9.77 \pm 0.3$	\\ 
CoDi	 & $0.505$	 & $0.482$	 & $0.429$	& $0.429$	 & $58.12 \pm 17.4$	\\ 
GReaT	 & $0.499$	 & $0.482$	 & $0.536$	& $0.589$	 & $9.80 \pm 0.3$	\\ 
CTGAN	 & $0.503$	 & $0.589$	 & $0.393$	& $0.482$	 & $47.72 \pm 9.7$	\\ 
TVAE	 & $N/A^*$	 & $0.036$	 & $0.089$	& $0.071$	 & $204.75 \pm 61.8$	\\ 
TabPFN	 & $0.499$ &$0.446$  &$0.536$  &$0.554$  & $38.44 \pm 4.1$	\\ 

\hline
    \end{tabular}}
    \end{subtable}    
\begin{flushleft}
\tiny
\textbf*{Note:} TVAE suffers from severe mode-collapse for the discrete data, collapsing to a single repeated row across all samples, causing MEC AUC to be incalculable using the framework and largely explains the poor performance for the other metrics. 
\end{flushleft}
\end{table*}

\begin{table*}[!t]
    \centering     
    \caption{20 nodes/variables Results \ref{tab:20-Low} Mean and standard deviation of metric values (error rate (\%) of single column density, error rate (\%) of pair-wise correlation score, $\alpha-$precision, $\beta-$recall) over $10$ random causal DAGs. \ref{tab:20-skel} Benchmark on causal skeletons. Values are mean and standard deviation of metric values
(SHD, F1 score, recall, and precision) over 10 random causal DAGs. Each metric value on a causal
graph is the average value over 5 bootstrapping datasets. \ref{tab:20-high} Causal direction level and Interventional downstream tasks. } 
    \label{tab:20}
    \begin{subtable}[b]{0.48\linewidth}
        \centering        
    \caption{Low-Order}
    \label{tab:20-Low}
    \resizebox{0.99\columnwidth}{!}{
    \begin{tabular}{l|llll}
    \hline
    & Col.  	 & Pair. 	 & $\alpha$-precision 	& $\beta$-recall \\ \hline 
TabSyn	 &$2.10 \pm 1.08$& $0.58 \pm 0.28$&$98.45 \pm 1.09$&$49.54 \pm 0.54$\\ 
STASY	 &$8.77 \pm 2.24$& $1.14 \pm 0.19$&$95.89 \pm 2.13$&$51.64 \pm 3.51$\\ 
TabDDPM	 &$0.99 \pm 0.21$& $1.77 \pm 1.13$&$98.13 \pm 0.53$&$49.84 \pm 0.88$\\ 
CoDi	 &$5.78 \pm 0.73$& $1.06 \pm 0.25$&$72.12 \pm 4.48$&$70.37 \pm 1.83$\\ 
GReaT	 &$8.60 \pm 0.47$& $1.29 \pm 0.99$&$75.89 \pm 5.88$&$58.70 \pm 2.17$\\ 
CTGAN	 &$5.42 \pm 0.84$& $3.55 \pm 0.52$&$81.14 \pm 4.65$&$10.49 \pm 6.56$\\ 
TVAE	 &$4.29 \pm 0.72$& $2.23 \pm 0.77$&$88.65 \pm 4.28$&$24.65 \pm 10.89$\\ 
TabPFN	 &$1.67 \pm 0.14$& $0.43 \pm 0.10$&$91.78 \pm 1.57$&$53.15 \pm 0.73$\\

\hline
    \end{tabular}}
    \end{subtable}   
    \begin{subtable}[b]{0.48\linewidth}
        \centering        
    \caption{Skel}
    \label{tab:20-skel}
    \resizebox{0.99\columnwidth}{!}{
    \begin{tabular}{l|llll}
    \hline
	& Adj  	 & F1 	 & Precision 	& Recall \\ \hline
ref.& $13.18\pm 3.90$	 & $0.83\pm 0.04$	 & $0.86\pm 0.09$	 & $0.82\pm 0.07$\\ 
\hline
TabSyn	 & $28.14\pm 6.67$	 & $0.68\pm 0.06$	 & $0.81\pm 0.10$	 & $0.60\pm 0.09$\\ 
STASY	 & $43.72\pm 8.64$	 & $0.58\pm 0.07$	 & $0.80\pm 0.10$	 & $0.46\pm 0.09$\\ 
TabDDPM	 & $56.14\pm 15.46$	 & $0.51\pm 0.09$	 & $0.78\pm 0.15$	 & $0.39\pm 0.07$\\ 
CoDi	 & $48.24\pm 14.64$	 & $0.55\pm 0.10$	 & $0.77\pm 0.09$	 & $0.44\pm 0.12$\\ 
GReaT	 & $31.64\pm 7.45$	 & $0.64\pm 0.06$	 & $0.77\pm 0.09$	 & $0.56\pm 0.06$\\ 
CTGAN	 & $102.64\pm 13.57$	 & $0.37\pm 0.05$	 & $0.83\pm 0.11$	 & $0.24\pm 0.03$\\ 
TVAE	 & $74.26\pm 15.61$	 & $0.46\pm 0.06$	 & $0.85\pm 0.11$	 & $0.32\pm 0.04$\\ 
TabPFN	 & $14.47\pm 4.89$	 & $0.81\pm 0.05$	 & $0.87\pm 0.07$	 & $0.77\pm 0.07$\\ 

\hline
    \end{tabular}}
    \end{subtable}    

    \begin{subtable}[b]{0.48\linewidth}
        \centering  
        \vspace{1em}
    \caption{High-Order}
    \label{tab:20-high}
    \resizebox{0.99\columnwidth}{!}{
    \begin{tabular}{l|lllll}
    \hline
	& MEC AUC  	 & RECI 	 & IGCI & CDS & Intv   \\ \hline
ref.& $0.980$	 & $0.454$	 & $0.378$& $0.487$ & $3.30 \pm 0.0$\\ 
\hline
TabSyn	 & $0.915$	 & $0.597$	 & $0.487$	& $0.496$	 & $4.66 \pm 1.7$	\\ 
STASY	 & $0.843$	 & $0.521$	 & $0.546$	& $0.496$	 & $19.51 \pm 5.5$	\\ 
TabDDPM	 & $0.720$	 & $0.513$	 & $0.597$	& $0.496$	 & $5.33 \pm 2.6$	\\ 
CoDi	 & $0.823$	 & $0.563$	 & $0.479$	& $0.538$	 & $7.01 \pm 2.2$	\\ 
GReaT	 & $0.812$	 & $0.395$	 & $0.597$	& $0.588$	 & $11.48 \pm 0.9$	\\ 
CTGAN	 & $0.550$	 & $0.630$	 & $0.471$	& $0.462$	 & $13.46 \pm 1.8$	\\ 
TVAE	 & $0.599$	 & $0.605$	 & $0.487$	& $0.546$	 & $9.04 \pm 2.8$	\\ 
TabPFN	 & $0.963$ & $0.455$ & $0.505$ & $0.525$ & $3.46 \pm 0.1$ \\ 

\hline
    \end{tabular}}
    \end{subtable}    
\end{table*}

\begin{table*}[!t]
    \centering     
    \caption{Imputation of the $P38$ feature for the Sachs dataset \cite{Sachs2005-su}. Metrics are low-order metrics from \cite{zhang2023mixed}, $r^2$ coefficient of determination, mean absolute error (MAE) and mean squared error (MSE). The evaluated models are XGBoost \cite{DBLP:journals/corr/ChenG16}, a simple MLP, and TabPFN \cite{tabpfn}.}\label{tab:Sachs}
    \begin{subtable}[b]{0.48\linewidth}
        \centering        
    \caption{Low-Order}
    % \label{tab:lg}
    \resizebox{0.99\columnwidth}{!}{
    \begin{tabular}{l|llll}
    \hline
        & Col.  	 & Pair. 	 & $\alpha$-precision 	& $\beta$-recall \\ \hline 
TabSyn	 &$2.54 \pm 0.51$& $2.30 \pm 0.71$&$98.17 \pm 0.96$&$48.54 \pm 1.06$\\ 
STASY	 &$12.54 \pm 4.72$& $3.08 \pm 1.43$&$91.78 \pm 2.97$&$48.25 \pm 7.58$\\ 
TabDDPM	 &$2.66 \pm 0.28$& $1.42 \pm 0.44$&$93.26 \pm 0.92$&$77.48 \pm 1.90$\\ 
CoDi	 &$37.47 \pm 3.57$& $10.03 \pm 2.04$&$57.39 \pm 4.89$&$32.21 \pm 4.69$\\ 
GReaT	 &$16.07 \pm 0.23$& $3.27 \pm 0.49$&$77.10 \pm 0.46$&$43.66 \pm 0.59$\\ 
CTGAN	 &$14.57 \pm 1.45$& $8.54 \pm 0.64$&$91.55 \pm 4.08$&$23.33 \pm 1.80$\\ 
TVAE	 &$13.39 \pm 3.00$& $5.69 \pm 1.31$&$85.22 \pm 3.22$&$36.46 \pm 1.93$\\ 
TabPFN	 &$4.52 \pm 0.21$& $8.17 \pm 0.73$&$94.66 \pm 0.89$&$50.46 \pm 0.48$\\ 

\hline
    \end{tabular}}
    \end{subtable}   
    \begin{subtable}[b]{0.49\linewidth}
        \centering        
    \caption{XGBoost}
    % \label{tab:lg}
    \resizebox{0.99\columnwidth}{!}{
    \begin{tabular}{l|lll}
    \hline
    & $r^2$  	 & MAE 	 &  MSE 	  \\ \hline 

TabSyn	 &$-0.02 \pm 1.81$ & $181.46 \pm 43.16$ & $2.38e+05 \pm 1.24e+05$\\ 
STASY	 &$0.87 \pm 0.08$ & $106.84 \pm 18.03$ & $7.89e+04 \pm 3.96e+04$\\ 
TabDDPM	 &$0.88 \pm 0.03$ & $108.56 \pm 11.35$ & $7.51e+04 \pm 1.82e+04$\\ 
CoDi	 &$0.57 \pm 0.31$ & $315.56 \pm 184.39$ & $4.96e+05 \pm 4.26e+05$\\ 
GReaT	 &$0.23 \pm 0.24$ & $275.58 \pm 43.68$ & $4.67e+05 \pm 1.35e+05$\\ 
CTGAN	 &$-22.57 \pm 27.91$ & $335.08 \pm 32.11$ & $7.35e+05 \pm 8.74e+04$\\ 
TVAE	 &$-10.98 \pm 12.03$ & $283.05 \pm 43.09$ & $6.42e+05 \pm 1.46e+05$\\ 
TabPFN   &$-5.05 \pm 5.29$&$261.23 \pm 79.17$ & $8.88e+05 \pm 7.00e+05$\\ 

\hline
    \end{tabular}}
    \end{subtable} 
    \\
    \vspace{1em}
    \begin{subtable}[b]{0.47\linewidth}
        \centering        
    \caption{MLP}
    % \label{tab:lg}
    \resizebox{0.98\columnwidth}{!}{
    \begin{tabular}{l|lll}
    \hline
    & $r^2$  	 & MAE 	 &  MSE 	  \\ \hline 
TabSyn   &$0.14 \pm 0.56$& $31.84 \pm 4.93$&$2.00e+04 \pm 2.09e+03$\\ 
STASY    &$0.47 \pm 0.08$& $28.12 \pm 2.96$&$1.78e+04 \pm 1.02e+03$\\ 
TabDDPM  &$0.47 \pm 0.06$& $28.04 \pm 1.32$&$1.70e+04 \pm 1.01e+03$\\ 
CoDi     &$0.45 \pm 0.08$& $46.89 \pm 16.63$&$2.36e+04 \pm 1.18e+04$\\ 
GReaT    &$-1.13 \pm 0.70$& $43.54 \pm 7.21$&$3.75e+04 \pm 5.86e+03$\\ 
CTGAN    &$-15.53 \pm 8.47$& $45.83 \pm 7.13$&$4.41e+04 \pm 7.97e+03$\\ 
TVAE     &$-4.29 \pm 8.27$& $28.11 \pm 1.80$&$2.93e+04 \pm 9.70e+03$\\ 
TabPFN   &$-2.38 \pm 1.19$& $72.51 \pm 9.86$&$6.23e+04 \pm 2.19e+04$\\

\hline
    \end{tabular}}
    \end{subtable}    
    \begin{subtable}[b]{0.5\linewidth}
        \centering        
    \caption{TabPFN}
    % \label{tab:lg}
    \resizebox{0.99\columnwidth}{!}{
    \begin{tabular}{l|lll}
    \hline
    & $r^2$  	 & MAE 	 &  MSE 	  \\ \hline 
TabSyn   &$-4.73 \pm 7.33$& $223.84 \pm 33.52$&$4.39e+05 \pm 1.51e+05$\\ 
STASY    &$-0.07 \pm 1.71$& $160.28 \pm 34.25$&$2.18e+05 \pm 1.15e+05$\\ 
TabDDPM  &$0.57 \pm 0.18$& $151.84 \pm 14.24$&$1.88e+05 \pm 4.08e+04$\\ 
CoDi     &$0.89 \pm 0.10$& $131.71 \pm 82.44$&$9.34e+04 \pm 8.30e+04$\\ 
GReaT    &$-1.07 \pm 0.96$& $217.61 \pm 22.92$&$3.39e+05 \pm 8.66e+04$\\ 
CTGAN    &$-3885.74 \pm 6312.95$& $363.49 \pm 29.36$&$8.71e+05 \pm 8.33e+04$\\ 
TVAE     &$-4.24 \pm 4.71$& $228.73 \pm 39.58$&$4.55e+05 \pm 1.03e+05$\\
TabPFN   &$-2.92 \pm 4.00$& $181.81 \pm 33.51$&$3.94e+05 \pm 1.79e+05$\\

\hline
    \end{tabular}}
    \end{subtable}    

\end{table*}

\subsection{More details and benchmarking results}
%% Comment about the table results for low-order evaluation
The results in Table~\ref{tab:low} show that CoDi stands out for its robust $\beta$-recall. However, this model tends to lag in $\alpha$-precision when compared to state-of-the-art results. On the other hand, TabDDPM and TabSyn achieve the lowest single column density estimation and pair-wise correlation errors. This suggests their ability to handle also the joint probability distributions between columns. Furthermore, STASY shows limitations in terms of single column density estimation task, having the highest error rates across all datasets and causal mechanisms.

\begin{table*}[t!]
    \centering     
    \caption{Benchmark on causal skeletons. The values represent  mean and standard deviation of metric values (SHD, F1 score, Recall, and Precision) over $10$ random causal DAGs. 
    Each metric value on a causal graph is averaged over $5$ bootstrapping datasets.}\label{tab:causal_skeleton_full}
    \begin{subtable}[b]{0.47\linewidth}
        \centering        
    \caption{Linear Gaussian }
    % \label{tab:lg}
    \resizebox{0.99\columnwidth}{!}{
    \begin{tabular}{l|llll}
    \hline
	& Adj  	 & F1 	 & Precision 	& Recall \\ \hline
ref.& $3.48\pm 1.69$	 & $0.90\pm 0.06$	 & $0.92\pm 0.07$	 & $0.89\pm 0.11$\\ 
\hline
TabSyn	& $12.96\pm 4.92$	 & $0.71\pm 0.12$	 & $0.88\pm 0.09$	 & $0.61\pm 0.16$\\ 
STASY	& $16.44\pm 9.19$	 & $0.68\pm 0.17$	 & $0.94\pm 0.07$	 & $0.56\pm 0.19$\\ 
TabDDPM	& $14.16\pm 8.28$	 & $0.70\pm 0.12$	 & $0.86\pm 0.11$	 & $0.61\pm 0.14$\\ 
CoDi	& $12.32\pm 3.29$	 & $0.72\pm 0.10$	 & $0.92\pm 0.09$	 & $0.60\pm 0.13$\\ 
GReaT	& $10.96\pm 3.53$	 & $0.75\pm 0.04$	 & $0.93\pm 0.08$	 & $0.64\pm 0.05$\\ 
CTGAN	& $35.04\pm 5.81$	 & $0.46\pm 0.06$	 & $0.88\pm 0.13$	 & $0.32\pm 0.06$\\ 
TVAE	& $23.32\pm 7.00$	 & $0.58\pm 0.06$	 & $0.89\pm 0.08$	 & $0.43\pm 0.07$\\ 
TabPFN	 & $4.28\pm 2.12$	 & $0.88\pm 0.07$	 & $0.92\pm 0.09$	 & $0.85\pm 0.09$\\ 
\hline
    \end{tabular}}
    \end{subtable}    
    \begin{subtable}[b]{0.49 \linewidth}
    \centering        
    \caption{Linear uniform}
    % \label{tab:lu}
    \resizebox{0.99\columnwidth}{!}{
    \begin{tabular}{l|llll}
    \hline
    		& Adj  	 & F1 	 & Precision 	& Recall \\ \hline
ref.	 & $4.04\pm 2.50$	 & $0.89\pm 0.07$	 & $0.91\pm 0.09$	 & $0.88\pm 0.10$\\ 
\hline
TabSyn	 & $13.16\pm 5.39$	 & $0.71\pm 0.08$	 & $0.89\pm 0.10$	 & $0.60\pm 0.10$\\ 
STASY	 & $14.80\pm 4.02$	 & $0.67\pm 0.10$	 & $0.89\pm 0.11$	 & $0.56\pm 0.12$\\ 
TabDDPM	 & $10.52\pm 5.37$	 & $0.77\pm 0.07$	 & $0.91\pm 0.09$	 & $0.67\pm 0.10$\\ 
CoDi	 & $11.12\pm 4.22$	 & $0.74\pm 0.09$	 & $0.93\pm 0.09$	 & $0.64\pm 0.11$\\ 
GReaT	 & $13.48\pm 5.70$	 & $0.70\pm 0.05$	 & $0.87\pm 0.09$	 & $0.60\pm 0.06$\\ 
CTGAN	 & $29.60\pm 6.69$	 & $0.51\pm 0.08$	 & $0.89\pm 0.13$	 & $0.37\pm 0.07$\\ 
TVAE	 & $25.28\pm 6.56$	 & $0.55\pm 0.07$	 & $0.89\pm 0.12$	 & $0.41\pm 0.06$\\ 
TabPFN	 & $4.84\pm 3.76$	 & $0.88\pm 0.06$	 & $0.92\pm 0.11$	 & $0.86\pm 0.07$\\ 
\hline
    \end{tabular}}
    \end{subtable}  
    \\
    \vspace{1em}
    \begin{subtable}[b]{0.48\linewidth}
        \centering        
    \caption{Sigmoid Gaussian (sample size: $1500$ )}
    \resizebox{0.99\columnwidth}{!}{
    \begin{tabular}{l|llll}
    \hline
    & Adj & F1 & Precision & Recall\\\hline
% real     & $7.76\pm 3.71$ & $0.80\pm 0.11$ & $0.94\pm 0.07$ & $0.73\pm 0.18$\\
% tabsyn   & $10.14\pm 4.18$ & $0.75\pm 0.12$ & $0.93\pm 0.06$ & $0.66\pm 0.18$\\
% STaSY    & $8.78\pm 3.85$ & $0.78\pm 0.12$ & $0.92\pm 0.08$ & $0.70\pm 0.17$\\
% tabddpm  & $13.30\pm 8.26$ & $0.64\pm 0.27$ & $0.76\pm 0.33$ & $0.59\pm 0.25$\\
% codi     & $8.46\pm 2.72$ & $0.79\pm 0.09$ & $0.94\pm 0.06$ & $0.70\pm 0.15$\\
% great    & $8.60\pm 3.69$ & $0.78\pm 0.11$ & $0.90\pm 0.08$ & $0.71\pm 0.16$\\
% ctgan    & $26.68\pm 5.78$ & $0.54\pm 0.09$ & $0.89\pm 0.09$ & $0.40\pm 0.10$\\
% tvae     & $13.30\pm 4.80$ & $0.71\pm 0.11$ & $0.94\pm 0.05$ & $0.60\pm 0.15$\\
ref. & $2.04\pm 1.46$	 & $0.95\pm 0.03$	 & $0.94\pm 0.06$	 & $0.95\pm 0.04$\\ 
\hline
TabSyn  & $3.00\pm 1.20$	 & $0.92\pm 0.02$	 & $0.96\pm 0.04$	 & $0.89\pm 0.05$\\ 
STASY  & $4.32\pm 2.12$	 & $0.88\pm 0.07$	 & $0.95\pm 0.06$	 & $0.84\pm 0.11$\\ 
TabDDPM  & $2.88\pm 1.26$	 & $0.92\pm 0.03$	 & $0.95\pm 0.06$	 & $0.91\pm 0.06$\\ 
CoDi  & $5.16\pm 2.29$	 & $0.87\pm 0.06$	 & $0.96\pm 0.05$	 & $0.80\pm 0.11$\\ 
GReaT  & $4.88\pm 3.21$	 & $0.87\pm 0.07$	 & $0.92\pm 0.10$	 & $0.84\pm 0.06$\\ 
CTGAN  & $20.40\pm 3.58$	 & $0.61\pm 0.05$	 & $0.94\pm 0.07$	 & $0.46\pm 0.06$\\ 
TVAE  & $11.12\pm 4.07$	 & $0.75\pm 0.08$	 & $0.96\pm 0.05$	 & $0.63\pm 0.11$\\ 
TabPFN	 & $2.70\pm 1.07$	 & $0.93\pm 0.02$	 & $0.94\pm 0.06$	 & $0.92\pm 0.04$\\ 
\hline
    \end{tabular}}
    \end{subtable}    
    \begin{subtable}[b]{0.48 \linewidth}
    \centering        
    \caption{Neural network Gaussian (sample size: $1500$)}
    \resizebox{0.99\columnwidth}{!}{
    \begin{tabular}{l|llll}
    \hline
    & Adj & F1 & Precision &Recall\\\hline
ref.& $5.28\pm 3.15$	 & $0.85\pm 0.07$	 & $0.81\pm 0.15$	 & $0.93\pm 0.07$\\ 
\hline
TabSyn & $6.12\pm 3.36$	 & $0.84\pm 0.06$	 & $0.89\pm 0.08$	 & $0.82\pm 0.09$\\ 
STASY & $6.32\pm 3.79$	 & $0.83\pm 0.07$	 & $0.81\pm 0.14$	 & $0.87\pm 0.07$\\ 
TabDDPM & $5.48\pm 3.21$	 & $0.85\pm 0.07$	 & $0.81\pm 0.14$	 & $0.91\pm 0.05$\\ 
CoDi & $6.68\pm 2.75$	 & $0.82\pm 0.05$	 & $0.86\pm 0.11$	 & $0.82\pm 0.09$\\ 
GReaT & $5.92\pm 2.28$	 & $0.84\pm 0.05$	 & $0.86\pm 0.09$	 & $0.84\pm 0.07$\\ 
CTGAN & $23.28\pm 5.63$	 & $0.58\pm 0.04$	 & $0.92\pm 0.08$	 & $0.43\pm 0.04$\\ 
TVAE & $14.72\pm 3.83$	 & $0.69\pm 0.08$	 & $0.93\pm 0.08$	 & $0.56\pm 0.09$\\ 
TabPFN	 & $5.42\pm 3.73$	 & $0.85\pm 0.09$	 & $0.80\pm 0.15$	 & $0.93\pm 0.05$\\ 

\hline
    \end{tabular}}
    \end{subtable}   
\end{table*}

\begin{table*}[t!]
    \centering     
    \caption{Benchmark on d-separations: Area under the curve scores (AUC) of ROC curves. sz represents the sample size.}\label{tab:dsep_full}
        \resizebox{0.9\linewidth}{!}{
    \begin{tabular}{l|l|lllllllll}
    \hline
     && ref. & TabSyn & STASY & TabDDPM & CoDi & GReaT& CTGAN & TVAE & TabPFN \\\hline
    \multirow{4}{*}{AUC}
    &LG (sz $15000$)&$0.972$&$0.927$&$0.930$&$0.814$&$0.917$&$0.921$&$0.566$&$0.703$&$0.970$\\ 
    &LU (sz $15000$)&$0.967$&$0.871$&$0.936$&$0.815$&$0.902$&$0.860$&$0.588$&$0.669$&$0.928$\\ 
    &SG (sz $5000$)&$0.982$&$0.974$&$0.963$&$0.982$&$0.956$&$0.964$&$0.559$&$0.826$&$0.977$\\ 
    &NN (sz $5000$)&$0.986$&$0.890$&$0.972$&$0.957$&$0.909$&$0.943$&$0.566$&$0.752$&$0.967$\\ 
    \hline
    \end{tabular}}
    \end{table*}

As for benchmarking on the causal skeleton and Markov equivalent class level, we use Fisher's Z-test and the kernel independence test for datasets with linear and nonlinear functional relationships, respectively.  
Because it is not feasible to apply kernel independence tests to datasets with large sample sizes.
Our experiments only include the results with small sample sizes, e.g., $1500$. 
Since such conditional independence tests are more reliable with larger sample sizes, the performance of baseline methods on the nonlinear datasets is less distinguishable and not as informative as the results on the linear datasets as shown in Table~\ref{tab:causal_skeleton_full} and Table~\ref{tab:dsep_full}.
As for the causal direction level, we apply three training-free bivariate causal discovery methods, RECI~\citep{blobaum2018cause}, IGCI~\citep{janzing2012information}, and CDS~\citep{fonollosa2019conditional}. We report the best of the three as the pairwise direction score in Table~\ref{tab:c_dir}. The aggregation reduces dependence on any single bivariate evaluator rather than implying one method dominates universally.

\begin{table*}[t!] 
\centering 
\caption{Benchmark on the causal direction level.} \label{tab:c_dir} 
\begin{subtable}[b]{0.48\linewidth}
\centering        
\caption{Evaluation with the accuracy ($\uparrow$) of recovering causal directions.}\label{tab:cdir_b}
\begin{tabular}{l|llll} \hline &LG & LU & SG  & NN \\ \hline 
 ref. &$0.50$&$1.00$&$1.00$&$0.98$\\ \hline 
 TabSyn &$0.50$&$0.95$&$\mathbf{1.00}$&$0.88$\\ 
 STASY &$\mathbf{0.70}$&$0.95$&$0.98$&$0.88$\\ 
 TabDDPM &$0.54$&$\mathbf{1.00}$&$0.89$&$0.86$\\ 
CoDi&$0.59$&$0.91$&$0.96$&$\mathbf{0.98}$\\ 
GReaT&$0.55$&$0.93$&$0.95$&$0.95$\\ 
 CTGAN &$0.55$&$0.82$&$0.88$&$0.88$\\ 
 TVAE &$0.54$&$0.84$&$0.89$&$0.89$\\ 
 TabPFN & $0.66$ & $0.96$ & $0.96$ & $0.91$ \\ 
 \hline 
\end{tabular} 
\end{subtable}
\hspace{2pt}
\begin{subtable}[b]{0.48\linewidth}
\centering        
    \caption{Evaluation with LiNGAM on linear uniform distribution data (bootstrapping times $5$). }\label{tab:lingam}    
    \begin{tabular}{l|ll}
    \hline
& SHD ($\downarrow$) 	 & F1 ($\uparrow$) 	 \\ 
\hline
ref.& $1.04 \pm 0.85$	 & $0.94 \pm 0.04$\\ 
\hline
TabSyn	& $21.66 \pm 5.75$	 & $0.41 \pm 0.06$\\ 
STASY	& $20.66 \pm 4.99$	 & $0.45 \pm 0.10$\\ 
TabDDPM	& $15.86 \pm 8.74$	 & $0.51 \pm 0.14$	 \\ 
CoDi	& $18.54 \pm 5.08$	 & $0.45 \pm 0.10$\\ 
GReaT	& $13.62 \pm 9.24$	 & $0.57 \pm 0.11$\\ 
CTGAN	& $33.64 \pm 5.02$ & $0.26 \pm 0.07$	\\ 
TVAE	& $26.62 \pm 8.15$ & $0.29 \pm 0.08$\\ 
TabPFN	 & $2.44 \pm 2.62$	 & $0.88 \pm 0.11$	\\ 
\hline
\end{tabular}
\end{subtable}
\end{table*}

\begin{table*}[t!]
\centering
\caption{Benchmark on causal discovery directionality using LiNGAM under two different data-generating assumptions with sample size $15000$ and bootstrapping $5$.}
\label{tab:causal_direction_lingam}
\begin{subtable}[b]{0.48\linewidth}
\centering
\caption{LiNGAM on linear \textbf{uniform} data.}
\resizebox{0.99\linewidth}{!}{
\begin{tabular}{l|llll}
\hline
& SHD & F1 & Precision & Recall \\
\hline
ref. & $1.04 \pm 0.85$ & $0.94 \pm 0.04$ & $0.98 \pm 0.03$ & $0.92 \pm 0.05$ \\
\hline
TabSyn & $21.66 \pm 5.75$ & $0.41 \pm 0.06$ & $0.85 \pm 0.09$ & $0.28 \pm 0.05$ \\
STASY & $20.66 \pm 4.99$ & $0.45 \pm 0.10$ & $0.96 \pm 0.10$ & $0.30 \pm 0.08$ \\
TabDDPM & $15.86 \pm 8.74$ & $0.51 \pm 0.14$ & $0.76 \pm 0.08$ & $0.39 \pm 0.14$ \\
CoDi & $18.54 \pm 5.08$ & $0.45 \pm 0.10$ & $0.84 \pm 0.15$ & $0.31 \pm 0.08$ \\
GReaT & $13.62 \pm 9.24$ & $0.57 \pm 0.11$ & $0.84 \pm 0.05$ & $0.44 \pm 0.11$ \\
CTGAN & $33.64 \pm 5.02$ & $0.26 \pm 0.07$ & $0.69 \pm 0.22$ & $0.16 \pm 0.04$ \\
TVAE & $26.62 \pm 8.15$ & $0.29 \pm 0.08$ & $0.63 \pm 0.18$ & $0.19 \pm 0.05$ \\
TabPFN	 & $2.44 \pm 2.62$	 & $0.88 \pm 0.11$	 & $0.95 \pm 0.08$	 & $0.83 \pm 0.13$\\ 
\hline
\end{tabular}}
\end{subtable}
\hfill
\begin{subtable}[b]{0.48\linewidth}
\centering
\caption{LiNGAM on linear \textbf{Gaussian} data.}
\resizebox{0.99\linewidth}{!}{
\begin{tabular}{l|llll}
\hline
& SHD & F1 & Precision & Recall \\
\hline
ref. & $15.42 \pm 7.01$ & $0.38 \pm 0.06$ & $0.49 \pm 0.05$ & $0.32 \pm 0.07$ \\
\hline
TabSyn & $27.74 \pm 5.14$ & $0.26 \pm 0.07$ & $0.51 \pm 0.17$ & $0.17 \pm 0.05$ \\
STASY & $31.92 \pm 4.55$ & $0.21 \pm 0.07$ & $0.45 \pm 0.13$ & $0.13 \pm 0.06$ \\
TabDDPM & $26.64 \pm 10.34$ & $0.24 \pm 0.09$ & $0.42 \pm 0.11$ & $0.18 \pm 0.08$ \\
CoDi & $29.66 \pm 5.12$ & $0.24 \pm 0.08$ & $0.50 \pm 0.13$ & $0.16 \pm 0.06$ \\
GReaT & $18.18 \pm 8.97$ & $0.36 \pm 0.06$ & $0.52 \pm 0.08$ & $0.28 \pm 0.07$ \\
CTGAN & $36.00 \pm 6.40$ & $0.20 \pm 0.06$ & $0.54 \pm 0.17$ & $0.13 \pm 0.04$ \\
TVAE & $29.60 \pm 8.47$ & $0.20 \pm 0.05$ & $0.42 \pm 0.12$ & $0.13 \pm 0.04$ \\
TabPFN	 & $14.66 \pm 6.16$	 & $0.38 \pm 0.07$	 & $0.48 \pm 0.09$	 & $0.33 \pm 0.07$\\ 
\hline
\end{tabular}}
\end{subtable}
\end{table*}
\begin{table*}[t!] 
\centering 
\caption{Benchmark on interventional and counterfactual tasks with sample size $1000$. Values are $100\times$ AMAEs (average mean absolute errors).}
\label{tab:eva_causal_inference}
\begin{subtable}[b]{0.48\linewidth}
    \centering
\caption{Intervention inference.}
\resizebox{0.99\columnwidth}{!}{
\begin{tabular}{l|llll}
\hline
& LG & LU & SG & NN \\
\hline
ref. & $3.16 \pm 0.2$ & $3.07 \pm 0.2$&  $3.3 \pm 0.1$& $3.3 \pm 0.2$\\
\hline
TabSyn & $4.73 \pm 1.8$ & $4.21 \pm 1.1$&  $4.6 \pm 1.1$& $4.7 \pm 0.8$\\
STASY & $26.66 \pm 7.1$ & $23.86 \pm 11.1$&  $25.7 \pm 10.5$& $21.3 \pm 5.8$\\
TabDDPM & $4.13 \pm 1.2$ & $3.48 \pm 0.6$&  $4.2 \pm 0.4$& $3.8 \pm 0.4$\\
CoDi & $5.25 \pm 1.6$ & $3.82 \pm 0.6$&  $10.2 \pm 4.2$& $9.5 \pm 3.9$\\
GReaT & $9.77 \pm 0.7$ & $11.20 \pm 1.2$&  $9.9 \pm 3.0$& $9.3 \pm 2.2$\\
CTGAN & $15.02 \pm 8.9$ & $10.89 \pm 3.2$&  $10.8 \pm 2.2$& $13.0 \pm 3.6$\\
TVAE & $9.52 \pm 5.2$ & $12.22 \pm 3.4$&  $7.1 \pm 1.3$& $9.4 \pm 2.7$\\
TabPFN & $3.35 \pm 0.3$ & $3.87 \pm 0.5$ & $3.95 \pm 0.4$ & $4.12 \pm 0.7$ \\
\hline
\end{tabular}}
\end{subtable}
\begin{subtable}[b]{0.46\linewidth}
\centering  
\caption{Counterfactual inference.}
\resizebox{0.99\columnwidth}{!}{
\begin{tabular}{l|llll}
\hline
& LG & LU & SG & NN \\ \hline
ref.& $0.04 \pm 0.0$ & $0.03 \pm 0.0$&  $0.32 \pm 0.2$&  $0.23 \pm 0.1$\\\hline
TabSyn& $0.56 \pm 0.4$ & $0.45 \pm 0.7$&  $0.88 \pm 0.4$&  $0.90 \pm 0.5$\\
STASY & $0.65 \pm 0.4$ & $0.44 \pm 0.3$&  $1.46 \pm 0.8$&  $1.63 \pm 0.9$\\
TabDDPM& $1.12 \pm 1.3$ & $0.46 \pm 0.7$&  $1.20 \pm 0.6$&  $0.75 \pm 0.3$\\
CoDi& $0.67 \pm 0.8$ & $0.58 \pm 0.4$&  $0.94 \pm 0.4$&  $1.53 \pm 0.7$\\
GReaT& $0.93 \pm 0.5$ & $0.58 \pm 0.4$&  $1.47 \pm 0.6$&  $2.60 \pm 1.4$\\
CTGAN& $8.58 \pm 7.7$ & $4.96 \pm 3.3$&  $5.02 \pm 2.2$&  $7.56 \pm 3.8$\\
TVAE& $5.22 \pm 3.9$& $5.40 \pm 3.6$&  $2.50 \pm 1.4$&  $4.53 \pm 2.5$\\
TabPFN & $0.21 \pm 0.1$ & $0.18 \pm 0.1$ & $0.56 \pm 0.2$ & $0.76 \pm 0.4$ \\
\hline
\end{tabular}}
\end{subtable}
\end{table*}
\subsection{More details and benchmarking results avoiding variable ordering bias}
\label{app:exp_details_var_order_bias}
\begin{table*}[!t]
    \centering     
    \caption{Benchmark on low-order statistics avoiding variable ordering bias. 
    Values are mean and standard deviation of metric values (error rate (\%) of single column density, error rate (\%) of pair-wise correlation score, $\alpha-$precision, $\beta-$recall) over $10$ random causal DAGs.}\label{tab:low_reordering}
    \begin{subtable}[b]{0.48\linewidth}
        \centering        
    \caption{Linear Gaussian}
    % \label{tab:lg}
    \resizebox{0.99\columnwidth}{!}{
    \begin{tabular}{l|llll}
    \hline
    & Col.  	 & Pair. 	 & $\alpha$-precision 	& $\beta$-recall \\ \hline 
TabSyn	 &$1.68 \pm 0.55$& $1.40 \pm 2.86$&$98.61 \pm 1.11$&$46.05 \pm 11.56$\\ 
STASY	 &$10.61 \pm 5.95$& $1.85 \pm 2.99$&$95.08 \pm 5.63$&$44.19 \pm 11.55$\\ 
TabDDPM	 &$0.82 \pm 0.15$& $0.52 \pm 0.38$&$99.10 \pm 0.65$&$50.32 \pm 0.38$\\ 
CoDi	 &$5.26 \pm 2.41$& $1.77 \pm 2.89$&$85.29 \pm 3.44$&$57.41 \pm 16.47$\\ 
GReaT	 &$8.74 \pm 0.27$& $0.99 \pm 0.39$&$80.43 \pm 6.34$&$52.67 \pm 1.16$\\ 
CTGAN	 &$5.67 \pm 0.57$& $4.17 \pm 0.91$&$86.89 \pm 5.62$&$14.75 \pm 10.47$\\ 
TVAE	 &$4.26 \pm 1.70$& $2.37 \pm 1.23$&$88.50 \pm 7.21$&$34.82 \pm 11.47$\\ 
TabPFN	 &$1.59 \pm 0.17$& $0.43 \pm 0.34$&$94.27 \pm 1.52$&$52.10 \pm 0.41$\\ 
\hline
    \end{tabular}}
    \end{subtable}    
    \begin{subtable}[b]{0.48 \linewidth}
    \centering        
    \caption{Linear uniform}
    % \label{tab:lu}
    \resizebox{0.99\columnwidth}{!}{
    \begin{tabular}{l|llll}
    \hline
    & Col.  	 & Pair. 	 & $\alpha$-precision 	& $\beta$-recall \\ \hline 
TabSyn	 &$1.73 \pm 0.92$& $0.45 \pm 0.27$&$98.78 \pm 0.78$&$49.61 \pm 1.55$\\ 
STASY	 &$8.82 \pm 3.25$& $1.17 \pm 0.55$&$95.72 \pm 2.90$&$48.71 \pm 2.58$\\ 
TabDDPM	 &$0.87 \pm 0.14$& $0.37 \pm 0.36$&$99.38 \pm 0.25$&$49.54 \pm 0.94$\\ 
CoDi	 &$9.19 \pm 2.52$& $1.65 \pm 1.42$&$65.09 \pm 4.18$&$61.27 \pm 4.16$\\ 
GReaT	 &$9.41 \pm 0.83$& $0.44 \pm 0.11$&$96.02 \pm 1.40$&$45.76 \pm 1.02$\\ 
CTGAN	 &$5.75 \pm 0.97$& $4.54 \pm 2.29$&$91.99 \pm 3.29$&$5.23 \pm 6.89$\\ 
TVAE	 &$4.98 \pm 1.52$& $2.35 \pm 0.75$&$94.34 \pm 2.35$&$10.43 \pm 10.45$\\ 
TabPFN	 &$1.82 \pm 0.62$& $0.35 \pm 0.12$&$98.73 \pm 0.54$&$50.44 \pm 0.76$\\ 
 
\hline
    \end{tabular}}
    \end{subtable}   
    \begin{subtable}[b]{0.48 \linewidth}
    \centering        
    \caption{Sigmoid Gaussian}
    % \label{tab:sg}
    \resizebox{0.99\columnwidth}{!}{
    \begin{tabular}{l|llll}
    \hline
    & Col.  	 & Pair. 	 & $\alpha$-precision 	& $\beta$-recall \\ \hline 
TabSyn	 &$1.62 \pm 0.60$& $0.60 \pm 0.33$&$98.64 \pm 0.82$&$49.55 \pm 0.52$\\ 
STASY	 &$10.06 \pm 2.57$& $1.71 \pm 0.83$&$93.42 \pm 3.53$&$46.18 \pm 5.20$\\ 
TabDDPM	 &$0.93 \pm 0.13$& $0.46 \pm 0.21$&$98.93 \pm 0.64$&$49.95 \pm 0.53$\\ 
CoDi	 &$7.46 \pm 2.43$& $1.58 \pm 0.90$&$91.96 \pm 2.73$&$65.53 \pm 3.02$\\ 
GReaT	 &$9.53 \pm 2.58$& $1.69 \pm 1.09$&$89.33 \pm 3.57$&$49.42 \pm 2.10$\\ 
CTGAN	 &$5.82 \pm 1.14$& $3.84 \pm 0.73$&$88.81 \pm 3.11$&$18.40 \pm 6.49$\\ 
TVAE	 &$4.65 \pm 0.67$& $1.64 \pm 0.74$&$92.06 \pm 3.14$&$37.86 \pm 6.24$\\ 
TabPFN	 &$1.91 \pm 0.16$& $0.48 \pm 0.14$&$95.60 \pm 0.89$&$51.73 \pm 0.66$\\ 
\hline
    \end{tabular}}
    \end{subtable}   
    \begin{subtable}[b]{0.48 \linewidth}
    \centering        
    \caption{Neural network Gaussian}
    % \label{tab:nn}
    \resizebox{0.99\columnwidth}{!}{
    \begin{tabular}{l|llll}
    \hline
    & Col.  	 & Pair. 	 & $\alpha$-precision 	& $\beta$-recall \\ \hline 
TabSyn	 &$2.46 \pm 2.26$& $3.08 \pm 7.96$&$98.47 \pm 1.72$&$44.87 \pm 14.91$\\ 
STASY	 &$7.69 \pm 2.34$& $3.98 \pm 8.11$&$96.15 \pm 3.12$&$43.47 \pm 14.71$\\ 
TabDDPM	 &$0.87 \pm 0.14$& $0.41 \pm 0.24$&$99.07 \pm 0.37$&$50.48 \pm 0.58$\\ 
CoDi	 &$7.02 \pm 2.19$& $3.53 \pm 7.68$&$86.73 \pm 2.74$&$54.64 \pm 18.31$\\ 
GReaT	 &$7.46 \pm 2.65$& $3.81 \pm 7.84$&$87.83 \pm 4.43$&$43.65 \pm 14.84$\\ 
CTGAN	 &$6.05 \pm 1.99$& $6.86 \pm 6.95$&$90.15 \pm 3.40$&$7.05 \pm 5.88$\\ 
TVAE	 &$5.53 \pm 1.59$& $4.20 \pm 7.34$&$90.26 \pm 4.45$&$20.56 \pm 12.09$\\ 
TabPFN	 &$1.60 \pm 0.32$& $0.47 \pm 0.08$&$95.36 \pm 1.94$&$51.40 \pm 0.53$\\ 

\hline
    \end{tabular}}
    \end{subtable}   
\vspace{2em}
\end{table*}
\begin{table*}[t!]
    \centering     
    \caption{Benchmark on causal skeletons avoiding variable ordering bias. 
    Values are mean and standard deviation of metric values (SHD, F1 score, Recall, and Precision) over $10$ random causal DAGs. 
    Each metric value on a causal graph is averaged over $10$ bootstrapping datasets.}\label{tab:causal_skeleton_full_reordering}
    \begin{subtable}[b]{0.48\linewidth}
        \centering        
    \caption{Linear Gaussian }
    % \label{tab:lg}
    \resizebox{0.99\columnwidth}{!}{
    \begin{tabular}{l|llll}
    \hline
	& Adj  	 & F1 	 & Precision 	& Recall \\ \hline
ref.	 & $4.04\pm 1.97$	 & $0.89\pm 0.05$	 & $0.91\pm 0.10$	 & $0.88\pm 0.08$\\ \hline
TabSyn	 & $14.66\pm 6.32$	 & $0.67\pm 0.17$	 & $0.82\pm 0.19$	 & $0.57\pm 0.17$\\ 
STASY	 & $17.62\pm 8.55$	 & $0.64\pm 0.17$	 & $0.86\pm 0.15$	 & $0.53\pm 0.19$\\ 
TabDDPM & $12.78\pm 6.24$	 & $0.72\pm 0.11$	 & $0.91\pm 0.09$	 & $0.62\pm 0.14$\\ 
CoDi	 & $15.12\pm 6.20$	 & $0.66\pm 0.17$	 & $0.86\pm 0.18$	 & $0.56\pm 0.17$\\ 
GReaT	 & $14.28\pm 6.75$	 & $0.69\pm 0.09$	 & $0.87\pm 0.13$	 & $0.59\pm 0.09$\\ 
CTGAN	 & $34.04\pm 5.87$	 & $0.48\pm 0.07$	 & $0.92\pm 0.10$	 & $0.33\pm 0.07$\\ 
TVAE	 & $23.52\pm 7.07$	 & $0.58\pm 0.06$	 & $0.90\pm 0.10$	 & $0.43\pm 0.06$\\ 
TabPFN	 & $7.92\pm 6.55$	 & $0.82\pm 0.10$	 & $0.89\pm 0.10$	 & $0.78\pm 0.12$\\ 

\hline
    \end{tabular}}
    \end{subtable}    
    \begin{subtable}[b]{0.48 \linewidth}
    \centering        
    \caption{Linear uniform}
    % \label{tab:lu}
    \resizebox{0.99\columnwidth}{!}{
    \begin{tabular}{l|llll}
    \hline
    		& Adj  	 & F1 	 & Precision 	& Recall \\ \hline
ref.	 & $4.24\pm 2.02$	 & $0.88\pm 0.04$	 & $0.90\pm 0.08$	 & $0.88\pm 0.08$\\ \hline
TabSyn	 & $17.02\pm 5.83$	 & $0.65\pm 0.09$	 & $0.88\pm 0.12$	 & $0.52\pm 0.10$\\ 
STASY	 & $18.14\pm 7.32$	 & $0.62\pm 0.15$	 & $0.86\pm 0.14$	 & $0.52\pm 0.17$\\ 
TabDDPM	 & $14.06\pm 6.52$	 & $0.69\pm 0.06$	 & $0.85\pm 0.12$	 & $0.60\pm 0.09$\\ 
CoDi	 & $12.14\pm 4.82$	 & $0.73\pm 0.10$	 & $0.91\pm 0.12$	 & $0.62\pm 0.13$\\ 
GReaT	 & $14.00\pm 4.42$	 & $0.69\pm 0.05$	 & $0.85\pm 0.09$	 & $0.58\pm 0.04$\\ 
CTGAN	 & $33.58\pm 8.31$	 & $0.50\pm 0.09$	 & $0.93\pm 0.07$	 & $0.34\pm 0.08$\\ 
TVAE	 & $27.28\pm 9.93$	 & $0.54\pm 0.08$	 & $0.88\pm 0.11$	 & $0.39\pm 0.08$\\ 
TabPFN	 & $7.82\pm 5.20$	 & $0.81\pm 0.09$	 & $0.89\pm 0.08$	 & $0.76\pm 0.12$\\ 
\hline
    \end{tabular}}
    \end{subtable}    
    \begin{subtable}[b]{0.48\linewidth}
        \centering        
    \caption{Sigmoid Gaussian (sample size: $500$ )}
    \resizebox{0.99\columnwidth}{!}{
    \begin{tabular}{l|llll}
    \hline
    & Adj & F1 & Precision & Recall\\\hline
ref.	 & $2.98\pm 2.35$	 & $0.91\pm 0.07$	 & $0.89\pm 0.08$	 & $0.94\pm 0.07$\\ \hline
TabSyn	 & $3.18\pm 2.35$	 & $0.91\pm 0.05$	 & $0.90\pm 0.07$	 & $0.93\pm 0.05$\\ 
STASY	 & $3.98\pm 2.75$	 & $0.88\pm 0.08$	 & $0.88\pm 0.10$	 & $0.90\pm 0.07$\\ 
TabDDPM	 & $3.96\pm 2.83$	 & $0.88\pm 0.08$	 & $0.87\pm 0.10$	 & $0.91\pm 0.07$\\ 
CoDi	 & $3.52\pm 2.37$	 & $0.89\pm 0.08$	 & $0.89\pm 0.08$	 & $0.91\pm 0.09$\\ 
GReaT	 & $5.10\pm 3.42$	 & $0.86\pm 0.08$	 & $0.85\pm 0.09$	 & $0.87\pm 0.09$\\ 
CTGAN	 & $15.32\pm 4.90$	 & $0.67\pm 0.04$	 & $0.87\pm 0.09$	 & $0.55\pm 0.04$\\ 
TVAE	 & $7.66\pm 2.59$	 & $0.80\pm 0.04$	 & $0.88\pm 0.08$	 & $0.75\pm 0.04$\\ 
TabPFN	 & $3.30\pm 2.61$	 & $0.90\pm 0.08$	 & $0.87\pm 0.10$	 & $0.94\pm 0.06$\\

\hline
    \end{tabular}}
    \end{subtable}    
    \begin{subtable}[b]{0.48 \linewidth}
    \centering        
    \caption{Neural network Gaussian (sample size: $500$)}
    \resizebox{0.99\columnwidth}{!}{
    \begin{tabular}{l|llll}
    \hline
    & Adj & F1 & Precision &Recall\\\hline
ref.	 & $6.00\pm 5.28$	 & $0.84\pm 0.13$	 & $0.78\pm 0.19$	 & $0.93\pm 0.05$\\ \hline
TabSyn	 & $6.92\pm 5.03$	 & $0.82\pm 0.10$	 & $0.80\pm 0.15$	 & $0.85\pm 0.07$\\ 
STASY	 & $6.20\pm 4.90$	 & $0.83\pm 0.11$	 & $0.78\pm 0.18$	 & $0.92\pm 0.06$\\ 
TabDDPM	 & $5.82\pm 5.20$	 & $0.84\pm 0.12$	 & $0.79\pm 0.18$	 & $0.93\pm 0.06$\\ 
CoDi	 & $5.84\pm 5.05$	 & $0.84\pm 0.12$	 & $0.79\pm 0.19$	 & $0.93\pm 0.04$\\ 
GReaT	 & $7.32\pm 4.49$	 & $0.79\pm 0.10$	 & $0.74\pm 0.15$	 & $0.88\pm 0.06$\\ 
CTGAN	 & $16.98\pm 7.08$	 & $0.64\pm 0.09$	 & $0.83\pm 0.16$	 & $0.53\pm 0.06$\\ 
TVAE	 & $11.48\pm 6.47$	 & $0.72\pm 0.12$	 & $0.79\pm 0.18$	 & $0.68\pm 0.09$\\ 
TabPFN	 & $8.10\pm 6.55$	 & $0.77\pm 0.14$	 & $0.69\pm 0.18$	 & $0.89\pm 0.11$\\ 

\hline
    \end{tabular}}
    \end{subtable}   
\end{table*}

\begin{table*}[t!]
    \centering     
    \caption{Benchmark on d-separations avoiding variables reordering: Area under the curve scores (AUC) of ROC curves. sz represents the sample size.}\label{tab:dsep_full_reordering}
        \resizebox{0.9\linewidth}{!}{
    \begin{tabular}{l|l|lllllllll}
    \hline
     && ref. & TabSyn & STASY & TabDDPM & CoDi & GReaT& CTGAN & TVAE & TabPFN \\\hline
    \multirow{4}{*}{AUC} &LG (sz=13500) & $0.966$&$0.840$&$0.844$&$0.864$&$0.819$&$0.844$&$0.528$&$0.658$&$0.897$\\ 
    & LU (sz=13500)&$0.972$&$0.749$&$0.857$&$0.759$&$0.839$&$0.808$&$0.536$&$0.547$&$0.917$\\ 
 
    &SG (sz=500)&$0.952$&$0.941$&$0.941$&$0.956$&$0.963$&$0.959$&$0.830$&$0.926$&$0.955$\\ 

    &NN (sz=500)&$0.962$&$0.940$&$0.954$&$0.965$&$0.944$&$0.924$&$0.788$&$0.860$&$0.916$\\

    \hline
    \end{tabular}}
    \end{table*} 
\begin{table*}[t!] 
\centering 
\caption{Benchmark on the causal direction level avoiding variable ordering bias: Evaluation with the accuracy ($\uparrow$) of recovering causal directions.} 
\label{tab:c_dir_reordering} 
\resizebox{0.9\textwidth}{!}{
\begin{tabular}{l|llll|llll|llll} 
\hline 
& \multicolumn{4}{c|}{RECI} & \multicolumn{4}{c|}{IGCI} & \multicolumn{4}{c}{CDS} \\ 
& LG & LU & SG & NN & LG & LU & SG & NN & LG & LU & SG & NN \\
\hline 
ref.    & 0.571 & 0.982 & 0.304 & 0.232 & 0.446 & 0.125 & 0.911 & 0.804 & 0.571 & 0.946 & 0.964 & 0.696 \\ \hline
TabSyn & 0.446 & 0.875 & 0.357 & 0.196 & 0.518 & 0.054 & 0.911 & 0.786 & 0.536 & 0.964 & 0.929 & 0.750 \\
STASY  & 0.464 & 0.964 & 0.321 & 0.232 & 0.536 & 0.089 & 0.857 & 0.804 & 0.500 & 0.857 & 0.946 & 0.786 \\
TabDDPM& 0.464 & 0.911 & 0.304 & 0.214 & 0.589 & 0.071 & 0.893 & 0.839 & 0.536 & 0.893 & 0.875 & 0.750 \\
CoDi   & 0.482 & 0.964 & 0.321 & 0.214 & 0.589 & 0.250 & 0.857 & 0.821 & 0.536 & 0.893 & 0.964 & 0.732 \\
GReaT  & 0.518 & 0.839 & 0.393 & 0.196 & 0.536 & 0.536 & 0.661 & 0.536 & 0.482 & 0.768 & 0.821 & 0.643 \\
CTGAN  & 0.643 & 0.768 & 0.393 & 0.214 & 0.625 & 0.286 & 0.804 & 0.786 & 0.571 & 0.607 & 0.250 & 0.268 \\
TVAE   & 0.661 & 0.714 & 0.339 & 0.232 & 0.571 & 0.179 & 0.893 & 0.857 & 0.500 & 0.839 & 0.464 & 0.321 \\
TabPFN & 0.464 & 0.875 & 0.393 & 0.268 & 0.482 & 0.161 & 0.857 & 0.893 & 0.482 & 0.911 & 0.964 & 0.554 \\

\hline 
\end{tabular}
}
\end{table*}

\begin{table*}[t!]
\centering
\caption{Benchmark on causal discovery directionality avoiding variables reordering using LiNGAM under two different data-generating assumptions with sample size $15000$ and bootstrapping $10$.}
\label{tab:causal_direction_lingam_reordering}
\begin{subtable}[b]{0.48\linewidth}
\centering
\caption{LiNGAM on linear \textbf{uniform} data.}
\resizebox{0.99\linewidth}{!}{
\begin{tabular}{l|llll}
\hline
& SHD & F1 & Precision & Recall \\
\hline
ref.	 & $1.54 \pm 1.58$	 & $0.91 \pm 0.09$	 & $0.94 \pm 0.08$	 & $0.89 \pm 0.09$\\ \hline
TabSyn	 & $27.58 \pm 6.32$	 & $0.34 \pm 0.09$	 & $0.79 \pm 0.17$	 & $0.22 \pm 0.07$\\ 
STASY	 & $23.81 \pm 3.74$	 & $0.40 \pm 0.11$	 & $0.88 \pm 0.11$	 & $0.26 \pm 0.09$\\ 
TabDDPM	 & $21.45 \pm 8.84$	 & $0.39 \pm 0.13$	 & $0.71 \pm 0.16$	 & $0.28 \pm 0.12$\\ 
CoDi	 & $21.57 \pm 4.93$	 & $0.37 \pm 0.11$	 & $0.73 \pm 0.18$	 & $0.25 \pm 0.09$\\ 
GReaT	 & $14.76 \pm 9.20$	 & $0.54 \pm 0.13$	 & $0.84 \pm 0.15$	 & $0.41 \pm 0.12$\\ 
CTGAN	 & $32.23 \pm 6.08$	 & $0.27 \pm 0.07$	 & $0.70 \pm 0.18$	 & $0.17 \pm 0.05$\\ 
TVAE	 & $26.62 \pm 10.30$	 & $0.34 \pm 0.11$	 & $0.72 \pm 0.19$	 & $0.22 \pm 0.08$\\ 
TabPFN	 & $8.20 \pm 6.35$	 & $0.68 \pm 0.20$	 & $0.83 \pm 0.20$	 & $0.59 \pm 0.19$\\ 
\hline
\end{tabular}}
\end{subtable}
\hfill
\begin{subtable}[b]{0.48\linewidth}
\centering
\caption{LiNGAM on linear \textbf{Gaussian} data.}
\resizebox{0.99\linewidth}{!}{
\begin{tabular}{l|llll}
\hline
& SHD & F1 & Precision & Recall \\
\hline
ref.	 & $15.80 \pm 7.15$	 & $0.36 \pm 0.05$	 & $0.45 \pm 0.04$	 & $0.31 \pm 0.06$\\ \hline
TabSyn	 & $23.40 \pm 7.53$	 & $0.31 \pm 0.08$	 & $0.54 \pm 0.16$	 & $0.21 \pm 0.07$\\ 
STASY	 & $29.95 \pm 4.87$	 & $0.24 \pm 0.08$	 & $0.51 \pm 0.13$	 & $0.16 \pm 0.06$\\ 
TabDDPM	 & $23.54 \pm 7.92$	 & $0.28 \pm 0.08$	 & $0.49 \pm 0.12$	 & $0.20 \pm 0.08$\\ 
CoDi	 & $30.23 \pm 5.38$	 & $0.20 \pm 0.06$	 & $0.41 \pm 0.11$	 & $0.13 \pm 0.04$\\ 
GReaT	 & $19.31 \pm 7.95$	 & $0.31 \pm 0.08$	 & $0.47 \pm 0.12$	 & $0.23 \pm 0.06$\\ 
CTGAN	 & $34.71 \pm 6.34$	 & $0.20 \pm 0.08$	 & $0.50 \pm 0.19$	 & $0.13 \pm 0.05$\\ 
TVAE	 & $27.87 \pm 8.31$	 & $0.23 \pm 0.09$	 & $0.46 \pm 0.19$	 & $0.15 \pm 0.07$\\ 
TabPFN	 & $18.02 \pm 9.96$	 & $0.32 \pm 0.12$	 & $0.41 \pm 0.13$	 & $0.27 \pm 0.12$\\ 
\hline
\end{tabular}}
\end{subtable}
\end{table*}
\begin{table*}[t!] 
\centering 
\caption{Benchmark on interventional and counterfactual tasks avoiding variable ordering bias with sample size $1000$. Values are $100\times$ AMAEs (average mean absolute errors).}
\label{tab:eva_causal_inference_reordering}
\begin{subtable}[b]{0.48\linewidth}
    \centering
\caption{Intervention inference.}
\resizebox{0.99\columnwidth}{!}{
\begin{tabular}{l|llll}
\hline
& LG & LU & SG & NN \\
\hline
ref.     & $3.19 \pm 0.2$       & $2.96 \pm 0.1$         & $3.21 \pm 0.2$       & $3.17 \pm 0.1$ \\
\hline
TabSyn   & $5.02 \pm 2.8$       & $4.19 \pm 1.4$         & $4.16 \pm 1.7$       & $9.08 \pm 14.6$ \\
STASY    & $24.66 \pm 12.3$     & $17.43 \pm 6.6$        & $23.29 \pm 6.4$      & $25.70 \pm 28.0$ \\
TabDDPM  & $3.87 \pm 1.2$       & $3.96 \pm 1.4$         & $3.62 \pm 0.5$       & $3.98 \pm 0.9$ \\
CoDi     & $9.14 \pm 10.8$      & $3.80 \pm 0.6$         & $16.20 \pm 7.5$      & $19.12 \pm 26.3$ \\
GReaT    & $11.50 \pm 2.3$      & $11.09 \pm 0.8$ & $12.96 \pm 3.2$   & $16.67 \pm 24.9$ \\
CTGAN    & $16.57 \pm 5.2$      & $15.19 \pm 7.9$        & $17.97 \pm 8.7$      & $26.12 \pm 27.2$ \\
TVAE     & $9.10 \pm 5.6$       & $10.70 \pm 3.8$        & $8.83 \pm 2.2$       & $22.44 \pm 36.8$ \\
TabPFN     & $3.81 \pm 1.7$ & $4.13 \pm 2.0$ & $4.24 \pm 0.5$ & $3.92 \pm 0.6$ \\
\hline
\end{tabular}}
\end{subtable}
\hfill
\begin{subtable}[b]{0.46\linewidth}
\centering  
\caption{Counterfactual inference.}
\resizebox{0.99\columnwidth}{!}{
\begin{tabular}{l|llll}
\hline
& LG & LU & SG & NN \\
\hline
ref.     & $0.05 \pm 0.0$ & $0.02 \pm 0.0$ & $0.07 \pm 0.0$ & $0.06 \pm 0.0$ \\
\hline
TabSyn   & $1.49 \pm 2.6$ & $0.52 \pm 0.8$ & $0.86 \pm 0.9$ & $10.74 \pm 30.7$ \\
STASY    & $2.31 \pm 3.4$ & $0.42 \pm 0.6$ & $3.87 \pm 2.4$ & $10.12 \pm 23.2$ \\
TabDDPM  & $0.96 \pm 1.4$ & $0.96 \pm 1.6$ & $0.65 \pm 0.4$ & $0.97 \pm 1.1$ \\
CoDi     & $2.57 \pm 4.4$ & $0.59 \pm 0.5$ & $4.74 \pm 4.0$ & $8.26 \pm 16.8$ \\
GReaT    & $1.49 \pm 1.7$ & $0.36 \pm 0.2$ & $3.24 \pm 1.9$ & $9.53 \pm 21.4$ \\
CTGAN    & $9.54 \pm 5.6$ & $7.92 \pm 5.4$ & $11.02 \pm 8.5$ & $18.69 \pm 25.9$ \\
TVAE     & $4.48 \pm 3.0$ & $5.14 \pm 3.2$ & $4.06 \pm 2.2$ & $13.16 \pm 22.3$ \\
TabPFN     & $1.01 \pm 2.4$ & $0.39 \pm 0.6$ & $0.78 \pm 0.4$ & $0.64 \pm 0.3$ \\
\hline
\end{tabular}}
\end{subtable}
\end{table*}

Tables~\ref{tab:low_reordering}, \ref{tab:causal_skeleton_full_reordering}, \ref{tab:c_dir_reordering}, 
\ref{tab:dsep_full_reordering}, \ref{tab:causal_direction_lingam_reordering}, and \ref{tab:eva_causal_inference_reordering} present a more detailed view of benchmarking results avoiding ordering bias. 
 To ensure unbiased modeling of causal information evaluation and avoid information leakage from data ordering as per the DAG, we randomly shuffle the variables of the benchmark data and train the synthetic tabular data model. This is done per seed such that the order of the benchmark dataset's columns is randomized, this ensures all synthesis models have consistent training data with a different order for every seed evaluted. This prevents the synthesis models from exploiting the original causal order. After synthetic data generation, we reverse the shuffling of the columns to restore the original order corresponding to the data generated from the DAG. This ensures the evaluation of high-order causal discovery metrics remains consistent.

\subsection{Benchmarking on real-world datasets}
\label{app:exp_realworld_datasets}
As for evaluating baseline methods on real-world data, there is no ground-truth causal DAG available $\mathcal{G}^{\texttt{gt}}$; hence, we consider the results of causal discovery methods on all the training data as pseudo labels. In this way, conclusions should be made carefully enough, because a worse performance compared with pseudo labels does not necessarily mean that the performance is poor but only represents the relative differences. As shown in Table~\ref{tab:real}, TabSyn is in general the best model over the four real-world datasets on causal skeleton-level evaluation. Though CoDi and GReaT can perform well on synthetic data, they do not outperform TabSyn in real-world datasets.

We pre-process the real-world dataset, Beijing, and remove the rows with any missing values and the date and time columns with strong correlation (almost deterministic relationship), "year", "month", "day", "hour", and "cbwd".

\begin{table*}[!t]
    \centering     
    \caption{Benchmark on low-order statistics on real-world datasets. 
    Values are mean and standard deviation of metric values (error rate (\%) of single column density, error rate (\%) of pair-wise correlation score, $\alpha-$precision, $\beta-$recall).}\label{tab:low-real}
    \begin{subtable}[b]{0.48\linewidth}
        \centering        
    \caption{Beijing}
    % \label{tab:beijing}
    \resizebox{0.99\columnwidth}{!}{
    \begin{tabular}{l|llll}
    \hline
	& Col.  	 & Pair. 	 & $\alpha$-precision 	& $\beta$-recall \\ \hline
TabSyn	 &$3.69$&$6.59$&$99.31$&$47.96$\\ 
STASY	 &$8.22$&$11.10$&$93.61$&$50.12$\\ 
TabDDPM	 &$63.50$*&$63.29$*&$0.55$*&$0.70$*\\ 
CoDi	 &$20.74$&$6.79$&$95.35$&$52.96$\\ 
GReaT	 &$9.57$&$60.92$&$97.36$&$60.19$\\ 
CTGAN	 &$19.37$&$24.89$&$96.38$&$39.48$\\ 
TVAE&$36.53$*&$40.97$*&$66.89$*&$1.69$*\\ 
\hline
    \end{tabular}}
    \end{subtable}    
    \begin{subtable}[b]{0.46 \linewidth}
    \centering        
    \caption{Magic}
    % \label{tab:magic}
    \resizebox{0.99\columnwidth}{!}{
    \begin{tabular}{l|llll}
    \hline
& Col.  	 & Pair. 	 & $\alpha$-precision 	& $\beta$-recall \\ \hline
TabSyn	 &$1.26$&$0.99$&$98.17$&$48.22$\\ 
STASY	 &$6.53$&$4.28$&$92.15$&$49.50$\\ 
TabDDPM	 &$0.79$&$1.33$&$98.66$&$47.32$\\ 
CoDi	 &$8.84$&$5.52$&$87.20$&$51.51$\\ 
GReaT	 &$15.16$&$9.66$&$85.17$&$39.90$\\ 
CTGAN	 &$4.43$&$7.33$&$90.29$&$15.13$\\ 
TVAE	 &$4.83$&$6.82$&$96.22$&$36.73$\\ 
\hline
    \end{tabular}}
    \end{subtable}   
    \begin{subtable}[b]{0.47 \linewidth}
    \centering        
    \caption{House}
    % \label{tab:house}
    \resizebox{0.99\columnwidth}{!}{
    \begin{tabular}{l|llll}
    \hline
& Col.  	 & Pair. 	 & $\alpha$-precision 	& $\beta$-recall \\ \hline
TabSyn	 &$3.76$&$1.60$&$95.53$&$40.41$\\ 
STASY	 &$8.27$&$2.46$&$96.82$&$49.31$\\ 
TabDDPM	 &$1.80$&$2.11$&$97.24$&$47.69$\\ 
CoDi	 &$26.09$&$5.69$&$77.07$&$34.57$\\ 
GReaT	 &$18.28$&$6.17$&$91.90$&$37.68$\\ 
CTGAN	 &$15.71$&$9.58$&$51.34$&$16.08$\\ 
TVAE	 &$10.77$&$4.72$&$95.37$&$26.62$\\ 
\hline
    \end{tabular}}
    \end{subtable}   
    \begin{subtable}[b]{0.48 \linewidth}
    \centering        
    \caption{Parkinsons}
    % \label{tab:parkinsons}
    \resizebox{0.99\columnwidth}{!}{
    \begin{tabular}{l|llll}
    \hline
& Col.  	 & Pair. 	 & $\alpha$-precision 	& $\beta$-recall \\ \hline
TabSyn	 &$1.47$&$22.63$&$95.08$&$27.43$\\ 
STASY	 &$28.07$&$24.84$&$60.39$&$21.11$\\ 
TabDDPM	 &$1.34$&$22.79$&$92.49$&$27.93$\\ 
CoDi	 &$11.57$&$26.61$&$91.66$&$38.86$\\ 
GReaT	 &$7.18$&$24.36$&$81.66$&$29.99$\\ 
CTGAN	 &$15.83$&$17.70$&$88.57$&$18.72$\\ 
TVAE	 &$7.66$&$6.55$&$88.86$&$33.10$\\ 
\hline
    \end{tabular}}
    \end{subtable}   

\end{table*}

\begin{table*}[t!]
    \centering     
    \caption{Benchmark on real-world data.}\label{tab:real}
    \begin{subtable}[b]{0.48 \linewidth}
    \centering        
    \caption{Beijing  (sample size: $15000$ )}
    % \label{tab:sg}
    \resizebox{0.99\columnwidth}{!}{
    \begin{tabular}{l|llll}
    \hline
	& SHD  	 &  F1 	 & Precision 	& Recall \\ \hline
ref.	 &  $1.20\pm 0.98$ 	 & $0.98\pm 0.02$ 	 & $1.00\pm 0.00$	 & $0.96\pm 0.03$\\ \hline
TabSyn	 &  $9.60\pm 1.74$ 	 & $0.80\pm 0.04$ 	 & $0.74\pm 0.04$	 & $0.88\pm 0.06$\\ 
STASY	 &  $\mathbf{2.40\pm 1.20}$ 	 & $\mathbf{0.96\pm 0.02}$ 	 & $1.00\pm 0.00$	 & $0.92\pm 0.04$ \\
TabDDPM &  $19.00\pm 1.00$ 	 & $0.71\pm 0.02$ 	 & $0.88\pm 0.04$	 & $0.59\pm 0.01$ \\
CoDi	 &  $13.40\pm 1.80$ 	 & $0.73\pm 0.03$ 	 & $0.69\pm 0.00$	 & $0.77\pm 0.06$ \\
GReaT	 &  $7.80\pm 1.40$ 	 & $0.84\pm 0.03$ 	 & $0.78\pm 0.04$	 & $0.91\pm 0.04$ \\
CTGAN	 &  $13.80\pm 1.89$ 	 & $0.76\pm 0.03$ 	 & $0.82\pm 0.04$	 & $0.70\pm 0.04$ \\
TVAE	 &  $21.80\pm 2.75$ 	 & $0.53\pm 0.06$ 	 & $0.47\pm 0.05$	 & $0.61\pm 0.07$ \\
\hline
    \end{tabular}}
    \end{subtable}   
    \begin{subtable}[b]{0.48 \linewidth}
    \centering        
    \caption{Magic (sample size: $15000$)}
    % \label{tab:nn}
    \resizebox{0.99\columnwidth}{!}{
    \begin{tabular}{l|llll}
    \hline
	& SHD  	 &  F1 	 & Precision 	& Recall \\ \hline
ref.	 &  $4.40\pm 3.56$ 	 & $0.94\pm 0.05$ 	 & $0.93\pm 0.05$	 & $0.95\pm 0.04$ \\\hline
TabSyn	 &  $\mathbf{9.40\pm 2.20}$ 	 & $\mathbf{0.88\pm 0.03}$ 	 & $0.86\pm 0.02$	 & $0.90\pm 0.04$ \\
STASY	 &  $12.20\pm 2.89$ 	 & $0.85\pm 0.04$ 	 & $0.83\pm 0.04$	 & $0.86\pm 0.05$ \\
TabDDPM	 & $17.00\pm 3.38$ 	 & $0.79\pm 0.03$ 	 & $0.82\pm 0.02$	 & $0.78\pm 0.05$  \\
CoDi	 &  $16.40\pm 3.32$ 	 & $0.81\pm 0.04$ 	 & $0.85\pm 0.03$	 & $0.77\pm 0.04$ \\
GReaT	 &  $20.40\pm 2.50$ 	 & $0.75\pm 0.03$ 	 & $0.77\pm 0.02$	 & $0.74\pm 0.04$ \\
CTGAN	 &  $18.80\pm 2.23$ 	 & $0.78\pm 0.02$ 	 & $0.86\pm 0.03$	 & $0.73\pm 0.03$ \\
TVAE	 &  $21.60\pm 3.44$ 	 & $0.73\pm 0.04$ 	 & $0.75\pm 0.06$	 & $0.72\pm 0.04$ \\
\hline
    \end{tabular}}
    \end{subtable}   
    \begin{subtable}[b]{0.48\linewidth}
        \centering        
    \caption{House (sample size: $15000$ )}
    % \label{tab:house}
    \resizebox{0.99\columnwidth}{!}{
    \begin{tabular}{l|llll}
    \hline
	& SHD  	 &  F1 	 & Precision 	& Recall \\ \hline
ref.	 &  $15.20\pm 3.25$ 	 & $0.90\pm 0.02$ 	 & $0.87\pm 0.03$	 & $0.92\pm 0.04$ \\\hline
TabSyn	 &  $30.80\pm 3.82$ 	 & $\mathbf{0.80\pm 0.02}$ 	 & $0.81\pm 0.03$	 & $0.79\pm 0.03$ \\
STASY	 &  $\mathbf{29.00\pm 3.82}$ 	 & $\mathbf{0.80\pm 0.03}$ 	 & $0.78\pm 0.04$	 & $0.83\pm 0.02$ \\
TabDDPM	 &  $51.40\pm 2.97$ 	 & $0.65\pm 0.02$ 	 & $0.63\pm 0.04$	 & $0.67\pm 0.02$ \\
CoDi	 &  $62.60\pm 3.58$ 	 & $0.61\pm 0.03$ 	 & $0.66\pm 0.04$	 & $0.58\pm 0.02$ \\
GReaT	 &  $64.40\pm 3.67$ 	 & $0.61\pm 0.02$ 	 & $0.66\pm 0.03$	 & $0.57\pm 0.02$ \\
CTGAN	 &  $80.80\pm 1.83$ 	 & $0.58\pm 0.01$ 	 & $0.73\pm 0.02$	 & $0.48\pm 0.01$ \\
TVAE	 &  $43.40\pm 5.73$ 	 & $0.72\pm 0.04$ 	 & $0.75\pm 0.04$	 & $0.70\pm 0.04$ \\
\hline
    \end{tabular}}
    \end{subtable}    
    \begin{subtable}[b]{0.48 \linewidth}
    \centering        
    \caption{Parkinsons (sample size: $5000$ )}
    % \label{tab:lu}
    \resizebox{0.99\columnwidth}{!}{
    \begin{tabular}{l|llll}
    \hline
	& SHD  	 &  F1 	 & Precision 	& Recall \\ \hline
ref.	 &  $4.80\pm 3.37$ 	 & $0.91\pm 0.06$ 	 & $0.99\pm 0.03$	 & $0.84\pm 0.11$ \\\hline
TabSyn	 &  $\mathbf{6.60\pm 2.20}$ 	 & $\mathbf{0.86\pm 0.04}$ 	 & $0.95\pm 0.04$	 & $0.80\pm 0.07$ \\
STASY	 &  $19.60\pm 1.74$ 	 & $0.62\pm 0.02$ 	 & $0.72\pm 0.03$	 & $0.54\pm 0.03$ \\
TabDDPM	 &  $8.20\pm 2.60$ 	 & $0.84\pm 0.04$ 	 & $0.98\pm 0.04$	 & $0.74\pm 0.06$ \\
CoDi	 &  $14.80\pm 2.04$ 	 & $0.72\pm 0.03$ 	 & $0.88\pm 0.04$	 & $0.62\pm 0.04$ \\
GReaT	 &  $13.00\pm 1.84$ 	 & $0.75\pm 0.03$ 	 & $0.88\pm 0.04$	 & $0.65\pm 0.04$ \\
CTGAN	 &  $33.20\pm 3.37$ 	 & $0.46\pm 0.05$ 	 & $0.65\pm 0.07$	 & $0.36\pm 0.04$ \\
TVAE	 &  $22.00\pm 1.26$ 	 & $0.61\pm 0.02$ 	 & $0.79\pm 0.04$	 & $0.50\pm 0.02$ \\
\hline
    \end{tabular}}
    \end{subtable}   
\end{table*}

\section{Use of Large Language Models (LLMs)}
In preparing this submission, LLMs were used solely as general-purpose writing assistants. Their role was limited to grammar checking, minor rephrasing for clarity, and ensuring consistency of style. LLMs were not involved in research ideation, experimental design, data analysis, or the generation of novel scientific content. All conceptual contributions, methodological developments, experiments, and results are entirely the work of the authors.

The authors take full responsibility for the content of this paper.

\clearpage

\end{document}